\documentclass[sigconf, authorversion]{acmart}
\usepackage{multirow}
\usepackage{makecell}
\usepackage{colortbl}
\usepackage{enumitem}
\usepackage{bbding}
\usepackage{wrapfig}
\usepackage{balance}
\AtBeginDocument{%
  \providecommand\BibTeX{{%
    \normalfont B\kern-0.5em{\scshape i\kern-0.25em b}\kern-0.8em\TeX}}}

\acmYear{2023} 
\setcopyright{acmlicensed}
\acmConference[MM '23]{Proceedings of the 31st
ACM International Conference on Multimedia}{October 29-November 3,
2023}{Ottawa, ON, Canada}
\acmBooktitle{Proceedings of the 31st ACM International Conference on
Multimedia (MM '23), October 29-November 3, 2023, Ottawa, ON, Canada}
\acmPrice{15.00}
\acmDOI{10.1145/3581783.3612148}
\acmISBN{979-8-4007-0108-5/23/10}

\begin{document}

\title{Question-Answer Cross Language Image Matching for Weakly Supervised Semantic Segmentation}

\author{Songhe Deng}
\email{dengsonghe2022@email.szu.edu.cn}
\affiliation{%
  \institution{Computer Vision Institute, School of Computer Science and Software Engineering, Shenzhen University}
}

\author{Wei Zhuo}
\email{weizhuo@szu.edu.cn}
\affiliation{%
  \institution{National Engineering Laboratory for Big Data System Computing Technology, Shenzhen University}
  }

\author{Jinheng Xie}
\email{xiejinheng2020@email.szu.edu.cn}
\affiliation{%
  \institution{Computer Vision Institute, School of Computer Science and Software Engineering, Shenzhen University}
  }

\author{Linlin Shen}
\authornotemark[1]
\email{llshen@szu.edu.cn}
\affiliation{%
  \institution{Computer Vision Institute, School of Computer Science and Software Engineering, Shenzhen University}
  }


\begin{abstract}

Class Activation Map (CAM) has emerged as a popular tool for weakly supervised semantic segmentation (WSSS), allowing the localization of object regions in an image using only image-level labels.
However, existing CAM methods suffer from under-activation of target object regions and false-activation of background regions due to the fact that a lack of detailed supervision can hinder the model's ability to understand the image as a whole.
In this paper, we propose a novel Question-Answer Cross-Language-Image Matching framework for WSSS (QA-CLIMS), leveraging the vision-language foundation model to maximize the text-based understanding of images and guide the generation of activation maps.
First, a series of carefully designed questions are posed to the VQA (Visual Question Answering) model with Question-Answer Prompt Engineering (QAPE) to generate a corpus of both foreground target objects and backgrounds that are adaptive to query images.
We then employ contrastive learning in a Region Image Text Contrastive (RITC) network to compare the obtained foreground and background regions with the generated corpus.
Our approach exploits the rich textual information from the open vocabulary as additional supervision, enabling the model to generate high-quality CAMs with a more complete object region and reduce false-activation of background regions.
We conduct extensive analysis to validate the proposed method and show that our approach performs state-of-the-art on both PASCAL VOC 2012 and MS COCO datasets.
Code is available at: \textcolor{magenta}{\textit{https://github.com/CVI-SZU/QA-CLIMS}}
\end{abstract}

\ccsdesc[500]{Computing methodologies~Image segmentation}
\keywords{Weakly-Supervised Learning, Semantic Segmentation, Visual Question Answering, Vision-Language Learning}

\maketitle

\section{Introduction}

Semantic segmentation is an important and challenging task in computer vision. Commonly, it requires pixel-wise semantic labels to train a neural network in a fully-supervised manner. However, annotating pixel labels is an exhaustive work that is both time and labor consuming. 
To alleviate the workload, we consider learning the segmentation model in a weakly-supervised manner that requires image-level labels only.

The current approaches for weakly supervised semantic segmentation (WSSS) typically train a classification network first to generate class activation maps (CAMs) as initial pseudo labels. 
These pseudo labels are then refined during the subsequent stage, ultimately resulting in fully supervised training based on the refined pseudo labels.
However, since CAMs activate not only the target object but also contextual information that aids in class discrimination, they often result in activation confusion between the target object and non-target objects or things that frequently co-occur~\cite{lee2022wood,xie2022clims,lee2021railroad, Xie_2021_ICCV}. For instance, as introduced in CLIMS~\cite{xie2022clims}, the model may mistakenly activate the ``tracks'' as ``train''. In some cases, CAMs even overemphasize background contextual information over the target objects, exacerbating object under-activation.

To address these challenges, previous works~\cite{jiang2022l2g, lee2021railroad, lee2021reducing} often employ an additional saliency map
to supervise the training of CAMs.
However, the saliency method is class-agnostic and cannot completely prevent background activation. 
A later work~\cite{lee2022wood} trained a classifier with specifically collected background images. This classifier is then used to differentiate background and foreground representations. In another research direction, recent works~\cite{xie2022clims, lin2022clipes} leverage the recent success of visual-large scale language pre-training(VLP) for weakly-supervised semantic segmentation. Different from previous methods that are based on images purely, 
these methods take advantage of easily accessible text corpus and as well the general knowledge of VLP learned on open-set data. In particular, they introduce the cross language-image matching to refine the CAMs, and meanwhile, manually collect text corpus about background objects and things to tackle the aforementioned challenge of incorrect background activation. The manually collected text corpus, namely negative corpus set, however, is typically limited by the knowledge of annotators and unable to be tailored to specific images. A fixed negative set cannot cover all possible cases, and it will also  introduce additional bias.

\begin{figure}[t]
  \centering
  \includegraphics[width=0.95\linewidth]{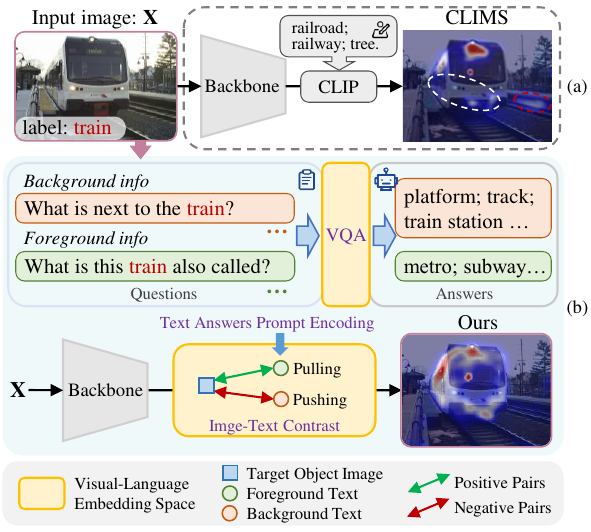}
  \vspace{-0.4cm}
  \caption{(a) CLIMS. Using predefined class-related background text labels.
  (b) Our QA-CLIMS designed a question-answer prompt engineering procedure to enhance the text-based understanding of image and refine CAMs through contrastive learning on both the foreground target, backgrounds and their corresponding texts.
  }
  \Description{None}
  \label{fig:intro}
\vspace{-0.7cm}
\end{figure}

To this end, we propose a novel Question-Answer Cross-Langu-age-Image Matching framework for WSSS (QA-CLIMS). This framework is built on the recent success of vision-Language Pre-training (VLP) and its text decoder to guide the query-adaptive training of CAM generation. Different from previous methods, we proposed a novel Question-Answer Prompt Engineering (QAPE) procedure and a Region Image-Text Contrastive(RITC) network that models more comprehensive language-image relations. Instead of using a fixed negative corpus set, we make full use of the image-grounded text decoder of the VLP foundation models, and propose our QAPE procedure to generate text prompts for both the foreground target object and the backgrounds that are adaptive to query images. The query-adaptive text prompts are experimentally proved more efficient for background suppression and activation refinement. Based on the proposed text prompts, our RITC network models comprehensive contrastive relations between foregrounds and backgrounds through the cross language image matching. Note that, The VLP has been widely used as an accessible function model for many subsequent tasks. With our designs on the VLP models, we provide a query-adaptive weakly-supervised semantic segmentation pipeline which is supposed to be general in applications.

For our framework, we first proposed a question-answer prompt engineering procedure to generate a query-adaptive corpus about the scene of the input query image. Inspired by the recent VLP models, we make full use of its image-grounded text decoder, i.e. the Visual Question Answering (VQA) function of VLP foundation models, to acquire global semantics about the input scene. How to get informative and helpful answers from the VQA model is, however, not easy. Question design and its subsequent engineering on the noisy answers are both crucial for success. To this end, we design intensive experiments on various types of questions and get a set of useful questions about both foreground and background, and we then post-process the answers with ground truth image labels to eliminate semantic ambiguity. 
Specifically, we ask a VQA model a set of pre-defined questions and get answers about the input image. We found that our framework can benefit more from the high-level questions about global semantics instead of the questions about image low-level features such as color and texture. 
Additionally, we found aliases for foreground objects are still useful. It could be due to the fact that open vocabulary doesn't necessarily follow one language habit.
More details and valuable experience can be found in our method and experimental sections. Since our approach employs automatically generated corpus rather than a fixed negative corpus set, it takes the advantage of low human involvement and cost, as well as confusion elimination in salient activation.

Given the text prompts, we build a region image-text contrastive (RITC) network, which consists of a backbone for CAM generation and an image-text contrastive supervision module. 
During this stage, we fully leverage the alignment attribute between the image encoder and text encoder provided by the VLP model. 
We encourage a matching between a masked image of the object region and its corresponding text, such as an image of a masked train and its category "train".
We suppose that the VLP module can recognize the input masked image very well if the predicted mask of the target object is accurate, and its encoding can surely be matched to the corresponding text prompts as well. 
Specifically, we mainly design a foreground region contrastive (FRC) loss and a background region contrastive (BRC) loss. They enforce matching of the image-text pair belonging to the same class and facilitate contrast learning of encodings when the text and image are from different classes.
To build effective training, we proposed a novel foreground adaptive thresholding module that thresholds masks to obtain the foreground and background region images, instead of directly using the activation map.
This training mechanism effectively enhances the quality of class activation maps.

We summarize our contribution as follows:
\begin{itemize}[leftmargin=0.6cm]
\item We proposed a general QA-CLIMS framework that maximizes the text-based understanding of
images (e.g., target object aliases, text label of surrounding object and
scene) for weakly-supervised semantic segmentation. 
\item We propose a novel QAPE procedure, that can generate query-adaptive text prompts, and a RITC network that models more comprehensive contrastive relations than previous works.
\item Experimental results show that our proposed QA-CLIMS achieves state-of-the-art performance with 3.4\% improvement on PASCAL VOC 2012 and 2.5\% gain on MS COCO 2014 dataset.
\end{itemize}

\section{Related work}

\subsection{Weakly Supervised Semantic Segmentation}

Weakly-Supervised Semantic Segmentation (WSSS) gains great interest all the time due to its great application values in reality. The target of WSSS is to learn the recognition of pixels through annotations which requires less annotating workload compared to pixel labels, such as image labels \cite{chang2020sccam, chen2022recam, zhang2021adaptive, zhang2022multi, Xie_2021_ICCV}, bounding boxes \cite{pu2018graphnet, khoreva2017simple}, and scribbles \cite{lin2016scribblesup, vernaza2017learning}. Image labels, which is a type of annotation most easily acquired in reality, are as well used and studied most widely. Early works that use only image labels generate Class Activation Maps (CAMs) as pseudo-labels for training.
CAMs usually incorrectly activate the background or surrounding non-target objects. Later work~\cite{jiang2022l2g, lee2021railroad, lee2021reducing, xu2022boatinsky} attempted to tackle this issue by generating saliency maps as additional supervision. 
Saliency maps, however, are class-agnostic and have the same issue of wrong activation as well. In addition, the saliency detector usually performs inferior on non-prominent objects and small objects. W-OoD \cite{lee2022wood} proposed to use extra out-of-distribution data to suppress the spurious correlations in WSSS, but the process of collecting data requires manual intervention and is inefficient. Differently, C2AM~\cite{C2AM} proposed a contrastive learning based on foreground-background discrimination, which can also reduce the over-activation of class-related background.

Recently, some text-based methods have been proposed for WSSS task and achieved competitive performance. 
CLIMS \cite{xie2022clims} introduced the image-text matching model
to refine the initial CAMs.
CLIP-ES \cite{lin2022clipes} generates CAMs directly from CLIP by matching with text labels. 
Typically, these methods use predefiend background label text, which is difficult to cover all surrounding obejcts and backgrounds.

\subsection{Vision-language Pre-training}

Vision-language pre-training (VLP) \cite{CLIP, li2022blip, wang2021simvlm} is a technique that aims to jointly learn visual and textual representations from large-scale multimodal data, such model has received tremendous success on various multimodal downstream tasks.
Previous works \cite{CLIP, jia2021ALIGN, li2021declip} typically train an image encoder and a text encoder, which allows computation of the similarity between images and text.
Considering the domain differences between masked images and natural images, Liang \textit{et al.} \cite{liang2022ovseg} fine-tuned CLIP on the collected region-word pairs, adapting it to semantic segmentation task.
However, such encoder-based model can only perform the understanding of images, which limits the model's capability.
Some recent works \cite{li2022blip, wang2021simvlm, chen2022pali} has allowed the text generation task by adding a text decoder in the model.
For example, BLIP \cite{li2022blip} treats visual question answering (VQA) as an answer generation problem and shows good performances in diverse image domains, for a large variety of real-world objects.

\subsection{Contrastive Learning}

Contrastive learning methods \cite{kaiming2020moco, chen2020simclr, CLIP} have shown great success and potential in various computer vision tasks, which focus on mining higher quality feature representations by contrasting positive and negative samples. 
InfoNCE loss \cite{oord2018infonce} is widely used to pull positive samples closer and push other negative examples away. 
SimCLR \cite{chen2020simclr} proposed to use different data augmentation methods to form positive and negative samples for each image instance. 
CLIP \cite{CLIP} collected a large dataset of 400 million image-text pairs for image-language contrastive learning. For each image, they treated the corresponding caption as positive sample and the rest of captions in the same batch as negative samples. 
Different from these, in our work, we leverage image-text contrastive learning for our novel designed question-answer weakly-supervised semantic segmentation framework, focusing on distinguishing target objects from surrounding objects and backgrounds.

\begin{figure*}[t]
  \centering
  \includegraphics[width=\textwidth]{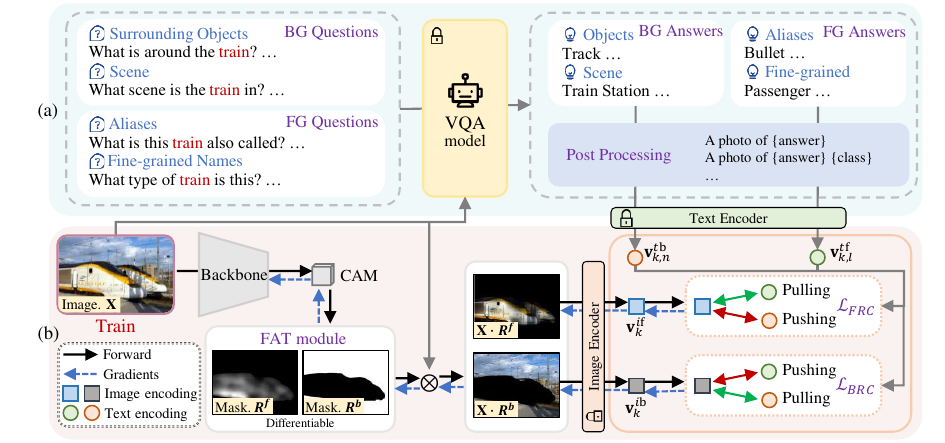}
  \vspace{-0.8cm}
  \caption{An overview of the proposed QA-CLIMS. Our framework consists of two parts: (a) The Question-Answer Prompt Engineering (QAPE) and (b) the Region Image-Text Contrastive (RITC) network. During the QAPE procedure, we generate a set of image-related semantic texts with pre-defined FG and BG questions and a VQA model. After post-processing the generated texts, we then train the RITC network for CAM. The network is equipped with our foreground adaptive thresholding (FAT) module, and trained through our foreground region contrastive loss $\mathcal{L}_{FRC}$ and our background region contrastive loss $\mathcal{L}_{BRC}$. 
  }
  \Description{None}
  \label{fig:model}
\vspace{-0.3cm}
\end{figure*}

\section{Method}

\subsection{Overview}

Our proposed QA-CLIMS consists of two main parts: a Question-Answer Prompt Engineering (QAPE) module and a Region Image-Text Contrastive (RITC) network, as illustrated in Fig.\ref{fig:model}.
QAPE utilizes the ability of vision-language pre-training models to enhance the text-based understanding of images (e.g., text labels of surrounding objects and target object aliases).
Using a corpus constructed from rich textual information, we apply contrastive learning in the RITC network to optimize the generation of CAMs.

Given an input image $\mathrm{\textbf{X}}$ and its corresponding image-level label $k \in \mathbb{R}^{1\times K}$, our QAPE generates the class-related text labels of the target object and background things in the image. 
Subsequently, the initial CAM generated by the backbone network of RITC is imported to a foreground adaptive thresholding module to obtain foreground region and background region images. Without dense pixel labels, the backbone is initialized by a classification network that is trained roughly on image labels directly. We then import the text generated by QAPE and the region images to a language encoder and a visual encoder of a VLP foundation model, respectively. We treat the VLP foundation model as a great teacher and fully leverage its alignment function on image-text pairs. We encourage text-image matching on their encoding vectors when the text and image are both from the foreground or background, and push the encodings of the text and image away from each other if the text indicates the foreground object but the image contains the background, and vice versa.  Specifically, we employ three loss functions during training. These losses include the foreground region contrastive (FRC) loss, the background region contrastive (BRC) loss, and the area regularization (REG) loss. The total objective can be formulated as follows:
\begin{equation}\label{eqn-loss_overall}
\begin{aligned}
\mathcal{L} = \alpha \cdot \mathcal{L}_{FRC} + \beta \cdot \mathcal{L}_{BRC} + \gamma \cdot \mathcal{L}_{REG}.
\end{aligned}
\end{equation}
where $\alpha$, $\beta$ and $\gamma$ are adjustable loss weights. The designed objective enables our training to make full use of the rich textual information generated by QAPE, suppress false activation of background areas, and encourage full activation of foreground areas.

\subsection{Question-Answer Prompt Engineering}\label{section:method-understander}

The proposed Question-Answer Prompt Engineering (QAPE) is composed of a Foreground (FG) questions template set, a Background (BG) questions template set, and a Visual Question Answering(VQA) model $\mathcal{G}$ (BLIP \cite{li2022blip}), and a post processing step.
As previously noted, initial CAMs can produce false activation on surrounding objects and scenes, which are regarded as the background (BG) of the target objects. To address this issue, we designed a set of question templates guiding the model $\mathcal{G}$ to understand and generate the corresponding text labels.
Additionally, we consider the target object as foreground (FG) and designed a set of question templates to extract diverse labels for FG, thereby guiding CAM to activate more complete target regions.
The image and a series of questions are fed into model $\mathcal{G}$.
The resulting answers are post processed as the final text labels and applied to the RITC training.

\subsubsection{\textbf{Background Questions}}\label{section:bg_questions} 

We designed two sets of Background (BG) Questions to exploit information related to \textbf{Surrounding Objects} and \textbf{Scene}, respectively.
These questions encourage model $\mathcal{G}$ to explore non-target areas of the image to obtain a more comprehensive negative text corpus.

\textbf{Surrounding Objects.} 
As previously mentioned, CAMs can falsely activate objects located in the proximity of the target object, which are referred to as surrounding objects.
In order to identify these surrounding objects, we designed a set of question templates to acquire their names.
For instance, one such question template is "\textit{What is around the \{class\}?}", where "\textit{\{class\}}" is replaced by the text label of the target category. 
These generated questions are category-specific, thereby enabling the VQA model to focus on the area around the target object.
Our predefined question templates are designed to be generic for all categories.  
The full question templates can be found in Appendix.

\textbf{Scene.} 
For a complex scene, asking what is around the target may be not sufficient. The background image, i.e. an image with the target object masked out, actually depicts a scene where the target object lies in. Only one or two things in the background cannot represent the whole scene. Toward this observation, we designed a set of question templates to describe the scene from a global perspective. For instance, with the above questions asking surrounding objects, we get "\textit{glass}" as the background object for the target "\textit{person}". With  scene questions, such as "\textit{What scene is the \{class\} in?}", we get the answer "\textit{kitchen}", which is apparently a more accurate and comprehensive description of the background image. The above two kinds of questions are complementary: the answers to surrounding objects tell the segmentation model the exact object or thing to deactivate, while answers to the scene regularize the training with global semantic matching on image-text pairs.

\subsubsection{\textbf{Foreground Questions}}\label{section:fg_questions}

Using diverse text labels for each object can enhance the model's performance due to the variation in names, attributes, and characteristics of an object category. 
It is particularly useful because model $\mathcal{G}$ is trained on the open vocabulary and it learns the matching of an object to different expressions. 
With this observation, we divide Foreground (FG) Question templates into two sets: the questions about \textbf{Fine-grained Names} and \textbf{Object Aliases} respectively.

\textbf{Fine-grained Names.}
In order to obtain a more diverse textual representation of the target category in the image, we set up several question templates to achieve information about the object's fine-grained names.
For instance, we use question templates like "\textit{What type of \{class\} is this?}" to extract detailed information regarding the target category. With these questions, we can get various attributes that indicate a more fine-grained class the object belongs to. 

\textbf{Object Aliases.} 
To further enhance the textual representation of the target object, we define several questions, such as "\textit{What is this \{class\} also called?}", to extract additional aliases for the object.
An object may have different names in various countries and regions around the world, since English has a long and evolving history. To this end, these questions can help to capture these variations.

\subsubsection{\textbf{{Post processing}}}

We designed $N$ general background (BG) questions and $L$ general foreground (FG) questions in cloze format for all $k$ categories.
By filling in a class name into the questions templates, we get specific questions for each class.
We feed these two sets of questions into $\mathcal{G}$ with the image $\mathrm{\textbf{X}}$ to obtain the BG answers $\mathrm{\textbf{t}}_{k,n}^\mathrm{b}$ and the FG answers $\mathrm{\textbf{t}}_{k,l}^\mathrm{f}$.

It is not recommended to use these answer sets directly as they may contain some noise.
For example, the FG answer set may include adjectives or nouns that are related to the target object, without directly refering it.
To address this issue, we apply a post-processing step to optimize the produced answers.
Specifically, for the FG answer, we add the original category label at the end of the adjectives or nouns acquired from model $\mathcal{G}$.
For instance, the answer we get for the category "\textit{train}" is "\textit{passenger}", we then modify it to "\textit{passenger train}".
To provide original information, we augment the FG answers by appending a text of the original category label, resulting in a total of $L+1$ element in the FG answer set.
This approach provides a more complete and comprehensive representation of the target category than previous methods.
Finally, following the usual setup, we represent these optimized answers as "\textit{a photo of \{answer\}}", e.g., "\textit{a photo of passenger train}".

\subsection{Region Image-Text Contrastive Training}

After obtaining the class-related textual-based understanding of images, we design a Region Image-Text Contrastive (RITC) network to optimize the generation of CAMs by making full use of this information.

To start with, we use the backbone network to generate the initial CAMs.
We first extract the feature maps $\mathrm{\textbf{Z}} \in \mathbb R^{C\times H \times W}$ of input image $\mathrm{\textbf{X}}$, then apply a sigmoid function $\sigma$ to the feature maps 
and the learnable matrix $\mathrm{\textbf{W}} \in \mathbb{R}^{C\times K}$ in classification layer as follows:
\begin{equation}
\begin{aligned}
    \mathrm{\textbf{P}}_k (h,w) = \sigma (\mathrm{\textbf{W}}_k^{\top} \mathrm{\textbf{Z}} (h,w)).
\end{aligned}
\end{equation}
where $\mathrm{\textbf{P}}_k \in \mathbb{R}^{K \times H \times W}$ is the initial CAM of category $k$ and will be refined in the subsequent process.
We employed a Foreground Adaptive Thresholding Module to generate the foreground region mask $\mathrm{\textbf{R}}_k^\mathrm{f}$ and the background region mask $\mathrm{\textbf{R}}_k^\mathrm{b}$, which are elaborated in Sec.\ref{section:region-mask}.
By applying these masks to the image $\mathrm{\textbf{X}}$, we obtained the target object area and corresponding background area, respectively.
We train the network with cross image-text matching and our contrastive loss, which comprehensively models relations on the text-image pairs among the background and foreground objects.

\subsubsection{\textbf{Feature Encoding}}

In order to enable contrast of images and corresponding text, we employ the image encoder $f_i(\cdot)$ and text encoder $f_t(\cdot)$ from CLIP
that map the visual and textual inputs into a common feature space.
Specifically, we first obtain the representation vector $\mathrm{\textbf{v}}_k^{i\mathrm{f}}$ of foreground (FG) region image and $\mathrm{\textbf{v}}_k^{i\mathrm{b}}$ of background (BG) region image, respectively:
\begin{equation}\label{eqn-CLIP_v_feat}
\begin{aligned}
    &\mathrm{\textbf{v}}_k^{i\mathrm{f}} = f_i(\mathrm{\textbf{X}} \cdot \mathrm{\textbf{R}}_k^\mathrm{f}), \quad
    &\mathrm{\textbf{v}}_k^{i\mathrm{b}} = f_i(\mathrm{\textbf{X}} \cdot \mathrm{\textbf{R}}_k^\mathrm{b}).
\end{aligned}
\end{equation}

We then encode the text answers generated by QAPE above, obtaining the FG text representation $\mathrm{\textbf{v}}_{k,l}^{t\mathrm{f}}$ and the BG text representation $\mathrm{\textbf{v}}_{k,n}^{t\mathrm{b}}$ as follows:
\begin{equation}\label{eqn-CLIP_t_feat}
\begin{aligned}
    &\mathrm{\textbf{v}}_{k,l}^{t\mathrm{f}} = f_t(\mathrm{\textbf{t}}_{k,l}^\mathrm{f}), \quad
    &\mathrm{\textbf{v}}_{k,n}^{t\mathrm{b}} = f_t(\mathrm{\textbf{t}}_{k,n}^\mathrm{b}). 
\end{aligned}
\end{equation}

Furthermore, we computed the mean value of all FG text representation $\mathrm{\textbf{v}}_{k,l}^{t\mathrm{f}}$ and denoted it as $\overline{\mathrm{\textbf{v}}}_{k}^{t\mathrm{f}}$.
According to the paradigm of contrastive learning, we propose FRC and BRC loss to optimize the foreground region and background region in CAMs, respectively. 

\begin{table}[t]
  \caption{Comparison of the quality of generated CAMs and the pseudo ground-truth mask on {PASCAL VOC2012} \textit{train} set. Bac. denotes the backbone network for CAMs generation.}
  \label{tab:result-cam-voc}
  \vspace{-0.3cm}
  \begin{tabular}{lccc}
    \toprule
    Method & Bac. & CAM & Mask \\
    \midrule
    \ IRN \makecell[l]{\footnotesize CVPR'19} \cite{ahn2019irn}     &    R50        & 48.3  & 66.5   \\
    \ SC-CAM \makecell[l]{\footnotesize CVPR'20} \cite{chang2020sccam}   & WR38     & 50.9  & 63.4   \\
    \ AdvCAM \makecell[l]{\footnotesize CVPR'21} \cite{lee2021anti}   & R50     & 55.6  & 68.0   \\
    \ RIB \makecell[l]{\footnotesize NeurIPS'21} \cite{lee2021reducing} & R50    & 56.5 & 70.6 \\
    \ MCTformer \makecell[l]{\footnotesize CVPR'22} \cite{xu2022mctformer}   & ViT-B   & 61.7  & 69.1  \\
    \ CLIMS \makecell[l]{\footnotesize CVPR'22} \cite{xie2022clims}     & R50     & 56.6  & 70.5  \\ 
    \ W-OoD \makecell[l]{\footnotesize CVPR'22} \cite{lee2022wood}   & R50     & 59.1  & 72.1  \\
    \ AMN \makecell[l]{\footnotesize CVPR'22} \cite{lee2022amn}   & R50     & 62.1  & 72.2  \\
    \rowcolor{gray!20}
    \ QA-CLIMS (ours)  & R50   & \textbf{67.3} & \textbf{77.4} \\
  \bottomrule
\end{tabular}
\vspace{-0.5cm}
\end{table}

\subsubsection{\textbf{Foreground Region Contrastive (FRC) Learning}}

Given the $k$-th FG region representation $\mathrm{\textbf{v}}_k^{i\mathrm{f}}$, it should be similar to its FG text representation and dissimilar to the corresponding BG text representation. 
Thus, we treat $\overline{\mathrm{\textbf{v}}}_{k}^{t\mathrm{f}}$ as positive sample, while all $\mathrm{\textbf{v}}_{k,n}^{t\mathrm{b}}$ as negative samples. 
We first define the cosine similarity between the FG region representation and the corresponding text representation as follows:
\begin{equation}\label{eqn-sim_fg}
\begin{aligned}
s_{k}^{\mathrm{ff}} = \mathrm{sim}(\mathrm{\textbf{v}}_k^{i\mathrm{f}}, \overline{\mathrm{\textbf{v}}}_{k}^{t\mathrm{f}}), \quad
s_{k,n}^{\mathrm{fb}} = \mathrm{sim}(\mathrm{\textbf{v}}_k^{i\mathrm{f}}, \mathrm{\textbf{v}}_{k,n}^{t\mathrm{b}}).
\end{aligned}
\end{equation}

With similarity measured by cosine function, the contrastive loss function $\mathcal{L}_{FRC}$ is as follows:
\begin{equation}\label{eqn-loss_fg}
\begin{aligned}
\mathcal{L}_{FRC} = - \log \frac {\exp (s_{k}^{\mathrm{ff}} / \tau)} {\sum_{n=1}^N \exp (s_{k,n}^{\mathrm{fb}} / \tau) + \exp (s_{k}^{\mathrm{ff}}  / \tau)}, 
\end{aligned}
\end{equation}
where $\tau$ is a temperature hyper-parameter.
The generated initial CAMs will gradually approach the target object and further avoid false-activation of surrounding objects under the supervision of $\mathcal{L}_{FRC}$, which makes use of the rich BG text. 
However, the $\mathcal{L}_{FRC}$ alone does not encourage the model to explore non-discriminative object regions.

\subsubsection{\textbf{Background Region Contrastive (BRC) Learning}}

To make full use of the resulting text to improve the completeness of the activation object area, we design the loss function $\mathcal{L}_{BRC}$ to supervise the background region learning. 
Similarly, we have the cosine similarity of the BG region representation to FG and BG text:
\begin{equation}\label{eqn-sim_bg}
\begin{aligned}
s_{k}^{\mathrm{bf}} = \mathrm{sim}(\mathrm{\textbf{v}}_k^{i\mathrm{b}}, \overline{\mathrm{\textbf{v}}}_{k}^{t\mathrm{f}}), \quad
s_{k,n}^{\mathrm{bb}} = \mathrm{sim}(\mathrm{\textbf{v}}_k^{i\mathrm{b}}, \mathrm{\textbf{v}}_{k,n}^{t\mathrm{b}}).
\end{aligned}
\end{equation}

Given the $k$-th BG region representation, $\mathcal{L}_{BRC}$ let it approach the corresponding BG text representation more closely and move further away from its FG text representation:
\begin{equation}\label{eqn-loss_bg}
\begin{aligned}
\mathcal{L}_{BRC} = -\log \frac {\exp(\overline{s}_{k}^{\mathrm{bb}} / \tau) } {\exp(s_{k}^{\mathrm{bf}}) + \exp(\overline{s}_{k}^{\mathrm{bb}} / \tau))}, 
\end{aligned}
\end{equation}
where $\tau$ is the temperature hyper-parameter, $\overline{s}_{k}^{\mathrm{bb}}$ indicates the average of all $N$ similaries $s_{k,n}^{\mathrm{bb}}$.
When $\mathcal{L}_{BRC}$ is minimized, fewer target object pixels are reserved in BG regions and more target object contents are recovered in FG regions.

\subsubsection{\textbf{Foreground Adaptive Thresholding}}\label{section:region-mask}

As stated previously, we employ two distinct contrastive loss functions on the foreground and background regions of an image, respectively.
Current method typically directly multiply the activation map with original image to emphase the activated object and supress the backgrounds.
However, this approach may takes noise in foreground, which would affects the subsequent operations in negative way.
To this end, we introduce a foreground adaptive thresholding module that utilizes an adaptive threshold parameter $\theta_k$ to erase the background for foreground encoding with an adaptive threshold conditioned on the input.
Specifically, the threshold is determined using the highest confidence value in the initial CAM $\mathrm{\textbf{P}}_{k}$ and can be expressed as follows:
\begin{equation} \label{eqn-mask-theta}
    \begin{aligned}
        \theta_k = \omega \times \max(\mathrm{\textbf{P}}_k), \quad \omega \in [0,1], 
    \end{aligned}
\end{equation}
where $\max(\cdot)$ is the maximizing operation and $\omega$ is a ratio that controls the degree of filtering.
Subsequently, applying the threshold to the activation maps to obtain a binary mask $b_k$, where pixel values greater or equal to $\theta$ are set to 1 and those less than $\theta$ are set to 0.
Finally, obtain the foreground region mask $\mathrm{\textbf{R}}^{\mathrm{f}}_k$ and the background region mask $\mathrm{\textbf{R}}^{\mathrm{b}}_k$ as follows:
\begin{equation}
    \begin{aligned}
        \mathrm{\textbf{R}}^{\mathrm{f}}_k = \mathrm{\textbf{P}}_k \cdot b_k, \quad 
        \mathrm{\textbf{R}}^{\mathrm{b}}_k = (1-\mathrm{\textbf{P}}_k) \cdot (1-b_k)
    \end{aligned}
\end{equation}
Employing the proposed adaptive thresholding mechanism enables us to generate a more precise region mask for each image while mitigating the influence of noise on subsequent operations.

\begin{table}[t]
  \caption{Evaluation results on PASCAL VOC2012 \textit{val} and \textit{test} sets. Sup. denotes the type of supervision, including full ($\mathcal{F}$) supervision, image-level ($\mathcal{I}$) supervision, and the supervision with saliency maps ($\mathcal{S}$) and language ($\mathcal{L}$). 
  Seg. denotes the segmentation network. Bac. denotes the backbone network. V1: DeepLabV1. V2: DeepLabV2. R50: ResNet-50 \cite{he2016resnet}. $^\ddagger$: Segmentation network pretrained using MS COCO dataset.}
  \label{tab:result-seg-voc}
  \vspace{-0.3cm}
  \begin{tabular}{clcccc}
    \toprule
    Sup. & Method & Bac. & Seg. & Val & Test \\
    \midrule
    \ \multirow{2}*{$\mathcal{F}$} 
    & DeepLabV1 \makecell[l]{\footnotesize ICLR'15} \cite{chen2014deeplabv1} &   -  &  -   & 75.5  & -  \\
    \ & DeepLabV2 \makecell[l]{\footnotesize TPAMI'18} \cite{chen2017deeplabv2} &  -    & - & 77.6  & 79.7 \\
    \midrule
    \ \multirow{3}*{$\mathcal{I+S}$} 
    & RIB \makecell[l]{\footnotesize NeurIPS'21} \cite{lee2021reducing} &   R101  &  V2   & 70.2  & 70.0 \\
    & EPS \makecell[l]{\footnotesize CVPR'21} \cite{lee2021railroad} & R101 &  V2   & 70.9  & 70.8  \\
    & RCA \makecell[l]{\footnotesize CVPR'22} \cite{zhou2022regional}  &   R101  &  V2$^\ddagger$   & 72.2  & 72.8 \\
    \midrule
    \ \multirow{9}*{$\mathcal{I}$} 
    \ & IRN \makecell[l]{\footnotesize CVPR'19} \cite{ahn2019irn}      &   R50  &  V2   & 63.5  & 64.8  \\
    \ &SEAM \makecell[l]{\footnotesize CVPR'20} \cite{wang2020seam}  &  WR38     & V1 & 64.5  & 65.7 \\
    \ &SC-CAM \makecell[l]{\footnotesize CVPR'20} \cite{chang2020sccam}  &  R101     & V2$^\ddagger$ & 66.1  & 65.9 \\
    \ &CONTA \makecell[l]{\footnotesize NeurIPS'20} \cite{zhang2020conta} & WR38 & V1 & 66.1 & 66.7 \\
    \ &AdvCAM \makecell[l]{\footnotesize CVPR'21} \cite{lee2021anti}   &  R101   & V2  & 68.1  & 68.0  \\
    \ &W-OoD \makecell[l]{\footnotesize CVPR'22} \cite{lee2022wood} & WR38 &  V1 & 70.7  & 70.1 \\
    \ &AMN \makecell[l]{\footnotesize CVPR'22} \cite{lee2022amn} &  R101   & V2$^\ddagger$  & 70.7  & 70.6 \\
    \ &MCTformer \makecell[l]{\footnotesize CVPR'22} \cite{xu2022mctformer} & WR38 &  V1 & 71.9  & 71.6 \\
    \ &LPCAM \makecell[l]{\footnotesize CVPR'23} \cite{chen2023LPCAM} & R101 &  V2 & 70.1  & 70.4 \\
    \midrule
    \ \multirow{5}*{$\mathcal{I+L}$} 
    \ &CLIMS \makecell[l]{\footnotesize CVPR'22} \cite{xie2022clims}    & R101  &   V2   & 69.3  & 68.7   \\
    \ &CLIP-ES \makecell[l]{\footnotesize CVPR'23} \cite{lin2022clipes}    & R101  &   V2   & 71.1  & 71.4   \\
    \ &MMCST \makecell[l]{\footnotesize CVPR'23} \cite{xu2023MMCST}    & WR38  &   V1   & 72.2  & 72.2   \\
    \rowcolor{gray!20}
    \ &QA-CLIMS (ours) & R101 & V2 &  \textbf{72.4} & \textbf{72.3}\\
    \rowcolor{gray!40}
    \ &QA-CLIMS (ours) & WR38 & V1 &  \textbf{75.6} & \textbf{75.5}\\
  \bottomrule
\end{tabular}
\vspace{-0.3cm}
\end{table}

\subsubsection{\textbf{Area Regularization}}

We follow \cite{xie2022clims} perform a pixel-level area regularization term to constraint the size of activation maps:
\begin{equation}\label{eqn-loss_reg}
\begin{aligned}
\mathcal{L}_{REG} = \frac{1}{KHW} \sum_{k=1}^{K} \sum_{h=1}^{H} \sum_{w=1}^{W} \mathrm{\textbf{P}}_k(h,w).
\end{aligned}
\end{equation}

\begin{figure*}[t]
  \centering
  \includegraphics[width=\textwidth]{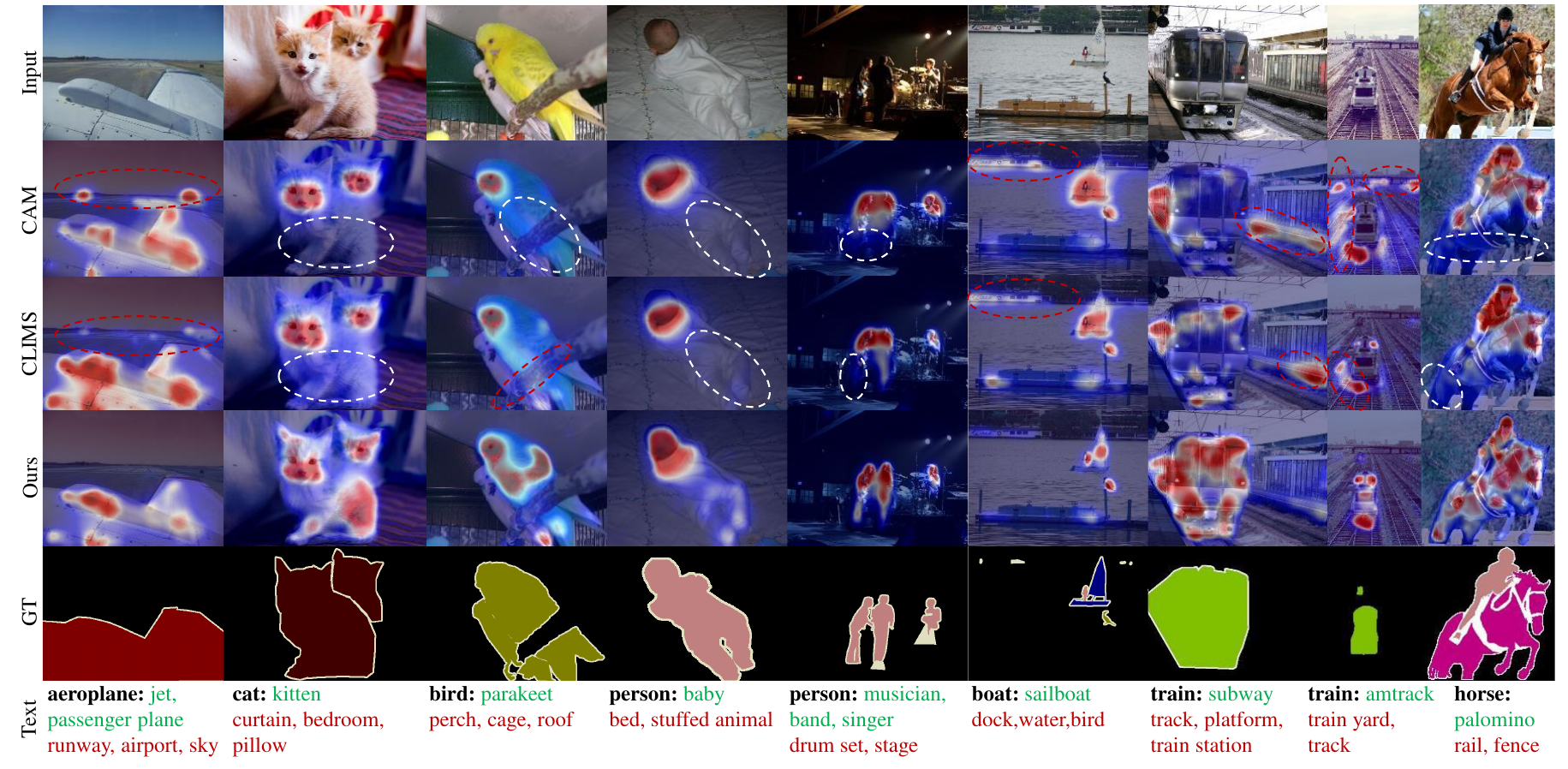}
  \vspace{-0.8cm}
  \caption{Visualization of the initial CAMs generated by CAM, CLIMS and the proposed QA-CLIMS. Last row is the text labels used in our QA-CLIMS, where the \textbf{bold} text represents the category label of target object, \textcolor[RGB]{75, 199, 131}{green} for the FG answer and \textcolor[RGB]{192, 0, 0}{red} for the BG answer. Best viewed in color.}
  \Description{None}
  \label{fig:cam}
\vspace{-0.3cm}
\end{figure*}

\begin{table}[t]
  \caption{Evaluation results on MS COCO 2014 validation set. Seg. denotes segmentation network.}
  \label{tab:result-seg-coco}
  \vspace{-0.3cm}
  \begin{tabular}{lccc}
    \toprule
    Method & Bac. & Seg. & Val  \\
    \midrule
    \ IRN \makecell[l]{\footnotesize CVPR'19} \cite{ahn2019irn}     &   R50  &  V2  & 32.6   \\
    \ SEAM \makecell[l]{\footnotesize CVPR'20} \cite{wang2020seam} & WR38 & V1 &  31.9 \\
    \ CONTA \makecell[l]{\footnotesize NeurIPS'20} \cite{zhang2020conta} & WR38  & V1  & 32.8 \\
    \ SIPE \makecell[l]{\footnotesize CVPR'22} \cite{chen2022sipe}   & R101   &  V2  & 40.6    \\
    \ URN \makecell[l]{\footnotesize AAAI'22} \cite{li2022urn}   & R101   &  V2  &   40.7  \\
    \ MCTformer  \makecell[l]{\footnotesize CVPR'22}  \cite{xu2022mctformer}    & WR38  &  V1   & 42.0    \\
    \rowcolor{gray!20}
    \ QA-CLIMS (ours) & R101 & V2 & \textbf{43.2}  \\
  \bottomrule
\end{tabular}
\vspace{-0.5cm}
\end{table}

\section{Experiments}

\subsection{Experimental Setup}

\textbf{Datasets and Evaluation Metric.} 
We evaluate the proposed method on PASCAL VOC 2012 \cite{everingham2009pascal} and MS COCO 2014 dataset \cite{lin2014microsoft}. 
VOC 2012 contains 20 object categories with 1464, 1449 and 1456 images in \textit{training}, \textit{validation}, and \textit{test} set, respectively. 
Following the common protocol in previous studies \cite{xie2022clims,ahn2019irn,lee2021anti}, VOC 2012 dataset was augmented to a set of 10582 images for training. 
COCO 2014 dataset contains 80 object classes with 82081 and 40137 images in \textit{training} and \textit{validation} set, respectively. 
In the training stage, we only use image labels. 
The mean Intersection over Union (mIoU) is adopted as the evaluation metric for all experiments.

\textbf{Implementation Details.} 
As indicated before, our method consists of two parts: a Question-Answer Prompt Engineering (QAPE) procedure and a Region Image-Text Contrastive (RITC) network.
For QAPE, we use BLIP \cite{li2022blip} as the VQA model.
In all experiments, we set 10 background (BG) question templates and 6 foreground (FG) question templates, which are provided in the Appendix.
Additionally, we use the image encoder and text encoder with the Mask-adapted technique \cite{liang2022ovseg} in our RITC.
We follow experimental settings of IRN \cite{ahn2019irn} to adopt ResNet-50 \cite{he2016resnet} as the backbone network for the generation of initial CAMs. 
SGD optimizer and poly-policy training schedule is adopted to train the model for 15 epochs, with the initial learning rate 3.5e-4. 
We set the filtering ratio $\omega$ to 0.1.
The temperature factors $\tau$ in both contrastive loss functions is 0.7.
For experiments on VOC dataset, we train our model on two GPUs using a batch size of 4 on each GPU, resulting in a total batch size of 8. 
The loss weights $\alpha$, $\beta$ and $\gamma$ are set to 10, 8 and 0.2, respectively.
For experiments on COCO, we set $\alpha=30$, $\beta=24$ and $\gamma=0.2$, while other settings remain the same. 
All models are implemented in PyTorch and trained on NVIDIA V100 GPU with 32 GB memory.

\textbf{Refinement of generated CAMs.} 
To obtain high-quality pseudo masks from the coarse CAMs, refinement techniques such as PSA \cite{ahn2018psa} and IRN \cite{ahn2019irn} that learn pixel-level semantic affinity with image-level supervision are commonly used. 
We follow existing work \cite{lee2022wood, chen2022recam, lee2021anti, lee2022amn, lee2021reducing} to adopt IRN for the CAM refinement.

\textbf{Semantic Segmentation.} 
Given pseudo ground-truth masks, we follow \cite{xie2022clims, lee2022amn, chen2022recam}  to adopt DeepLabV2 \cite{chen2017deeplabv2} with ResNet101 \cite{he2016resnet} backbone for the fully supervised segmentation network.
Moreover, to provide a more comprehensive comparison, we also utilized DeepLabV1 \cite{chen2014deeplabv1} with ResNet38 \cite{wu2019widerresnet} backbone, as used in \cite{wang2020seam, lee2022wood, chen2022semformer}.
All the backbones are pre-trained on ImageNet \cite{deng2009imagenet} dataset.

\begin{figure*}[t]
  \centering
  \includegraphics[width=\textwidth]{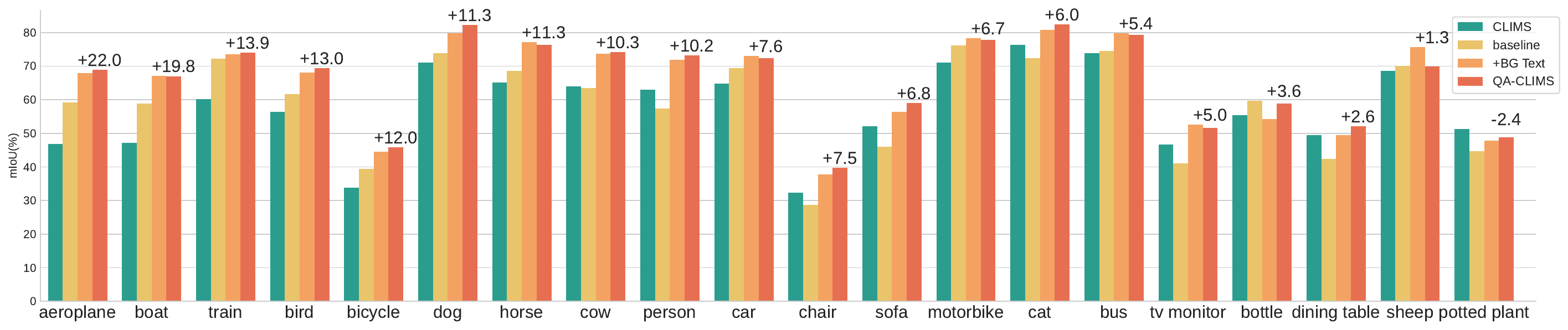}
  \vspace{-0.8cm}
  \caption{Per-class CAM quality comparison of different methods, including CLIMS, our baseline (using only FG text containing original category labels), using BG Text and our proposed QA-CLIMS.
  Evaluation is performed on PASCAL VOC 2012 \textit{train} set.
  The numbers above the bars represent the $\Delta$ 
  improvement (\%p) of QA-CLIMS compared to CLIMS.
  }
  \Description{None}
  \label{fig:compare-miou}
\vspace{-0.3cm}
\end{figure*}

\begin{table}[t]
  \caption{Compare the quality of initial CAMs on PASCAL VOC2012 \textit{train} set with different loss function combinations.}
  \label{tab:abexp-loss}
  \vspace{-0.3cm}
  \begin{tabular}{ccccc}
    \toprule
    $\mathcal{L}_{FRC}$ & $\mathcal{L}_{BRC}$ & $\mathcal{L}_{REG}$ & mIoU(\%) \\
    \midrule
    \ \Checkmark &  & &60.3\\
    \  & \Checkmark & & 57.7\\
    \ \Checkmark & \Checkmark & & 67.1 \\
    \rowcolor{gray!20}
    \ \Checkmark & \Checkmark & \Checkmark & \textbf{67.3}\\
  \bottomrule
\end{tabular}
\vspace{-0.5cm}
\end{table}

\subsection{Experimental Result}

\textbf{Quality of initial CAMs.} 
We first compare the performance of initial CAM and pseudo labels on PASCAL VOC \textit{training} set in Table \ref{tab:result-cam-voc}.
As shown in the table, the initial CAMs generated by our proposed method achieve 67.3\% mIoU, which outperforms numerous
recent state-of-the-art methods, e.g., MCTformer and AMN. 
We also compare the performance of the pseudo mask. As shown in the last row of Table \ref{tab:result-cam-voc}, our pseudo masks refined by IRN 
achieves the best mIoU of 77.4\%, which is 5.2\% higher than AMN.

In Fig.\ref{fig:cam}, we present the visualization of the generated CAMs and their corresponding text labels produced by QAPE.
Note that, we have not listed all the generated text labels due to space limitations.
we also compare the activation maps with those obtained from conventional CAM \cite{CAM} and CLIMS \cite{xie2022clims}, which is a method based on language image matching. 
It is important to note that text in Fig.\ref{fig:cam} is generated and utilized by our QA-CLIMS, while CLIMS employs a fixed collection of manually defined text.
The results demonstrate that QA-CLIMS typically activates more complete object contents while suppressing the surrounding background regions.
Specifically, as shown in the second to fourth columns, the CAM and CLIMS under-activate the cat, bird, and person regions, while QA-CLIMS produces more precise and comprehensive activation. The more visual comparison is shown in Fig.\ref{fig:cam}. 
We note that while CLIMS predefines a class-related background text set for certain object categories (e.g., railroad for train and river for boat categories), this predefined set of text cannot cover all possible cases.
For instance, as shown in the seventh column, although CLIMS avoids false activation of the railway, it still activates the platform area. 
Our proposed QAPE method, however, generates corresponding text labels for each image and is able to provide a more detailed and accurate description of the visual content. 
Guided by the $\mathcal{L}_{FRC}$ and $\mathcal{L}_{BRC}$ contrastive loss functions, our model is able to activate wider target regions and avoid activating the detailed background areas.

\textbf{Segmentation Performance.} 
With the pseudo labels as ground-truth, we perform traditional fully supervised learning to get the final segmentation model.
We present its semantic segmentation results on PASCAL VOC 2012 dataset in Table \ref{tab:result-seg-voc}. 
Our QA-CLIMS surpassing recent state-of-the-art methods on both \textit{validation} and \textit{test} sets, achieving 72.4\% and 72.3\% mIoU when using ResNet-101 backbone, and 75.6\% and 75.5\% mIoU with WideResNet-38 backbone.
Compared with MMCST \cite{xu2023MMCST}, our method outperforms it by 3.4\% and 3.3\% mIoU on \textit{validation} and \textit{test} sets, respectively.
We train the DeepLabV2 network on the MS COCO 2014 dataset using the generated pseudo labels and report the results in Table \ref{tab:result-seg-coco}. 
It is observed that our method achieves 43.2\% mIoU on \textit{validation} set, which is 2.5\% higher than URN \cite{li2022urn} and sets the best result.

\begin{table}[t]
  \caption{Comparison of the quality of generated CAMs using different combinations of text answers from questions. All results are achieved by using the $\mathcal{L}_{FRC}+\mathcal{L}_{BRC}+\mathcal{L}_{REG}$ loss.}
  \label{tab:abexp-question}
  \vspace{-0.3cm}
  \begin{tabular}{cccccc}
    \toprule
    \multicolumn{3}{c}{FG text} & \multicolumn{2}{c}{BG text} & \multirow{2}*{mIoU(\%)} \\
    \cmidrule(r){1-3} \cmidrule(r){4-5}
    Category & Fine-grained & Alias & Object & Scene &  \\
    \midrule
    \multicolumn{6}{l}{\textbf{RITC only.}} \\
    \Checkmark &     &    &   &   & 60.2\\
    \midrule
    \multicolumn{6}{l}{\textbf{+ corpus generated by QAPE.}} \\
    \Checkmark & \Checkmark &   &    &    & 61.0\\
    \Checkmark &   & \Checkmark &    &    & 60.7\\
    \Checkmark &   &   &  \Checkmark &    & 66.3\\
    \Checkmark &   &   &    &  \Checkmark & 64.5\\
    \Checkmark & \Checkmark & \Checkmark &    &    & 61.2\\
    \Checkmark & \Checkmark & \Checkmark &  \Checkmark &    & 67.0\\
    \rowcolor{gray!20}
    \Checkmark & \Checkmark & \Checkmark &  \Checkmark &  \Checkmark & \textbf{67.3}\\
  \bottomrule
\end{tabular}
\vspace{-0.5cm}
\end{table}

\subsection{Ablation and Analysis}

\textbf{Impact of Different Questions.}
As previously stated, a set of questions was designed to obtain text labels for different hypotheses.
To evaluate the effectiveness of different combinations of the obtained answers, we conducted experiments with various combinations of background (BG) text and foreground (FG) text.
The BG text was obtained from surrounding object-related and scene-related questions, while the FG text consisted of the original category label, the answers to the questions of fine-grained names and object aliases. 
The results of these experiments are presented in Table \ref{tab:abexp-question}. 
It is worth noting that when we train without QAPE procedure, we use only original category label as FG text and "\textit{no \{class\}}" as the BG text for contrast matching. 
For example, if the category label is "\textit{train}", the BG text will be "\textit{no train}".
The first row of the table shows that our RITC network results in a strong baseline of 60.2\%, while another text-based method CLIMS achieves a mIoU of 56.6\%. 
This result suggests that our improved contrastive loss and foreground adaptive thresholding module is effective in generating high-quality CAMs even without additional text data. 
Based on our baseline model, incorporating QAPE can further improve the overall performance.
Our findings show that Fine-grained Names are the most effective foreground questions while Surrounding Objects yield CAMs with the highest quality, among different background questions. Furthermore, combining answers from all question templates enhances the CAMs' quality. 
These results further justify the usefulness of each question template.

We also visualize the per-class variations when using different question combinations in Fig.\ref{fig:compare-miou} and we compare the results with CLIMS \cite{xie2022clims}.
It is observed that the CAMs of all categories, except the "\textit{potted plant}", can be greatly improved via our QA-CLIMS method. The BG texts contribute to the improvement of almost all the categories, while FG texts are of significant help for categories, such as "\textit{dog}" and "\textit{person}", which contain many fine-grained categories, like "\textit{terrier}" for "\textit{dog}", and "\textit{woman}", "\textit{child}" for "\textit{person}". 
The result proves that FG text provides more detailed and specific information about the object of interest, which can help the model to better focus on relevant features and suppress irrelevant ones.

\textbf{Impact of Loss Functions.} 
In this section, we evaluate the impact of each loss function in our method and summarizes the results in Table \ref{tab:abexp-loss}.
It is noteworthy that using only $\mathcal{L}_{FRC}$ can already achieve relatively high performance, and the combination of $\mathcal{L}_{FRC}$ and $\mathcal{L}_{BRC}$ can further improve the quality of CAM, resulting in a score of 67.1\%.
Surprisingly, we observe that using only $\mathcal{L}_{BRC}$ can also achieve a satisfactory mIoU score of 57.7\%.
Finally, incorporating the $\mathcal{L}_{REG}$ function produces the best outcomes.

\begin{wraptable}{r}{2.3cm}
  \vspace{-0.3cm}
  \begin{tabular}{cc}
    \toprule
    FAT & mIoU(\%) \\
    \midrule
    \  \XSolidBrush  & 66.9 \\
    \rowcolor{gray!20}
    \  \Checkmark  & \textbf{67.3}\\
  \bottomrule
\end{tabular}
\end{wraptable}
\textbf{Impact of Foreground Adaptive Thresholding (FAT).} 
Our analysis in right table demonstrates that our model can benefit from the improved masked images by the FAT module.

\section{Conclusion}

We propose a Question-Answer Cross-Language-Image Matching framework for WSSS (QA-CLIMS), which exploits the open-world perception capabilities of vision-language pretraining models to generate high-quality initial CAMs for weakly supervised semantic segmentation.
Experimental results demonstrate that our QA-CLIMS achieving state-of-the-art performance on PASCAL VOC2012 and MS COCO 2014 datasets.

\begin{acks}
This work was supported by the National Natural Science Foundation of China under Grant 82261138629; Guangdong Basic and Applied Basic Research Foundation under Grant 2023A1515010688 and Shenzhen Municipal Science and Technology Innovation Council under Grant JCYJ20220531101412030.
\end{acks}

\clearpage
\bibliographystyle{ACM-Reference-Format}
\balance
\bibliography{base}


\begin{thebibliography}{48}


\ifx \showCODEN    \undefined \def \showCODEN     #1{\unskip}     \fi
\ifx \showDOI      \undefined \def \showDOI       #1{#1}\fi
\ifx \showISBNx    \undefined \def \showISBNx     #1{\unskip}     \fi
\ifx \showISBNxiii \undefined \def \showISBNxiii  #1{\unskip}     \fi
\ifx \showISSN     \undefined \def \showISSN      #1{\unskip}     \fi
\ifx \showLCCN     \undefined \def \showLCCN      #1{\unskip}     \fi
\ifx \shownote     \undefined \def \shownote      #1{#1}          \fi
\ifx \showarticletitle \undefined \def \showarticletitle #1{#1}   \fi
\ifx \showURL      \undefined \def \showURL       {\relax}        \fi
\providecommand\bibfield[2]{#2}
\providecommand\bibinfo[2]{#2}
\providecommand\natexlab[1]{#1}
\providecommand\showeprint[2][]{arXiv:#2}

\bibitem[Ahn et~al\mbox{.}(2019)]%
        {ahn2019irn}
\bibfield{author}{\bibinfo{person}{Jiwoon Ahn}, \bibinfo{person}{Sunghyun Cho},
  {and} \bibinfo{person}{Suha Kwak}.} \bibinfo{year}{2019}\natexlab{}.
\newblock \showarticletitle{Weakly supervised learning of instance segmentation
  with inter-pixel relations}. In \bibinfo{booktitle}{\emph{Proceedings of the
  IEEE/CVF conference on computer vision and pattern recognition}}.
  \bibinfo{pages}{2209--2218}.
\newblock


\bibitem[Ahn and Kwak(2018)]%
        {ahn2018psa}
\bibfield{author}{\bibinfo{person}{Jiwoon Ahn} {and} \bibinfo{person}{Suha
  Kwak}.} \bibinfo{year}{2018}\natexlab{}.
\newblock \showarticletitle{Learning pixel-level semantic affinity with
  image-level supervision for weakly supervised semantic segmentation}. In
  \bibinfo{booktitle}{\emph{Proceedings of the IEEE conference on computer
  vision and pattern recognition}}. \bibinfo{pages}{4981--4990}.
\newblock


\bibitem[Chang et~al\mbox{.}(2020)]%
        {chang2020sccam}
\bibfield{author}{\bibinfo{person}{Yu-Ting Chang}, \bibinfo{person}{Qiaosong
  Wang}, \bibinfo{person}{Wei-Chih Hung}, \bibinfo{person}{Robinson Piramuthu},
  \bibinfo{person}{Yi-Hsuan Tsai}, {and} \bibinfo{person}{Ming-Hsuan Yang}.}
  \bibinfo{year}{2020}\natexlab{}.
\newblock \showarticletitle{Weakly-supervised semantic segmentation via
  sub-category exploration}. In \bibinfo{booktitle}{\emph{Proceedings of the
  IEEE/CVF Conference on Computer Vision and Pattern Recognition}}.
  \bibinfo{pages}{8991--9000}.
\newblock


\bibitem[Chen et~al\mbox{.}(2022d)]%
        {chen2022semformer}
\bibfield{author}{\bibinfo{person}{Junliang Chen}, \bibinfo{person}{Xiaodong
  Zhao}, \bibinfo{person}{Cheng Luo}, {and} \bibinfo{person}{Linlin Shen}.}
  \bibinfo{year}{2022}\natexlab{d}.
\newblock \showarticletitle{SemFormer: Semantic Guided Activation Transformer
  for Weakly Supervised Semantic Segmentation}.
\newblock \bibinfo{journal}{\emph{arXiv preprint arXiv:2210.14618}}
  (\bibinfo{year}{2022}).
\newblock


\bibitem[Chen et~al\mbox{.}(2014)]%
        {chen2014deeplabv1}
\bibfield{author}{\bibinfo{person}{Liang-Chieh Chen}, \bibinfo{person}{George
  Papandreou}, \bibinfo{person}{Iasonas Kokkinos}, \bibinfo{person}{Kevin
  Murphy}, {and} \bibinfo{person}{Alan~L Yuille}.}
  \bibinfo{year}{2014}\natexlab{}.
\newblock \showarticletitle{Semantic image segmentation with deep convolutional
  nets and fully connected crfs}.
\newblock \bibinfo{journal}{\emph{arXiv preprint arXiv:1412.7062}}
  (\bibinfo{year}{2014}).
\newblock


\bibitem[Chen et~al\mbox{.}(2017)]%
        {chen2017deeplabv2}
\bibfield{author}{\bibinfo{person}{Liang-Chieh Chen}, \bibinfo{person}{George
  Papandreou}, \bibinfo{person}{Iasonas Kokkinos}, \bibinfo{person}{Kevin
  Murphy}, {and} \bibinfo{person}{Alan~L Yuille}.}
  \bibinfo{year}{2017}\natexlab{}.
\newblock \showarticletitle{Deeplab: Semantic image segmentation with deep
  convolutional nets, atrous convolution, and fully connected crfs}.
\newblock \bibinfo{journal}{\emph{IEEE transactions on pattern analysis and
  machine intelligence}} \bibinfo{volume}{40}, \bibinfo{number}{4}
  (\bibinfo{year}{2017}), \bibinfo{pages}{834--848}.
\newblock


\bibitem[Chen et~al\mbox{.}(2022c)]%
        {chen2022sipe}
\bibfield{author}{\bibinfo{person}{Qi Chen}, \bibinfo{person}{Lingxiao Yang},
  \bibinfo{person}{Jian-Huang Lai}, {and} \bibinfo{person}{Xiaohua Xie}.}
  \bibinfo{year}{2022}\natexlab{c}.
\newblock \showarticletitle{Self-supervised image-specific prototype
  exploration for weakly supervised semantic segmentation}. In
  \bibinfo{booktitle}{\emph{Proceedings of the IEEE/CVF Conference on Computer
  Vision and Pattern Recognition}}. \bibinfo{pages}{4288--4298}.
\newblock


\bibitem[Chen et~al\mbox{.}(2020)]%
        {chen2020simclr}
\bibfield{author}{\bibinfo{person}{Ting Chen}, \bibinfo{person}{Simon
  Kornblith}, \bibinfo{person}{Mohammad Norouzi}, {and}
  \bibinfo{person}{Geoffrey Hinton}.} \bibinfo{year}{2020}\natexlab{}.
\newblock \showarticletitle{A simple framework for contrastive learning of
  visual representations}. In \bibinfo{booktitle}{\emph{International
  conference on machine learning}}. PMLR, \bibinfo{pages}{1597--1607}.
\newblock


\bibitem[Chen et~al\mbox{.}(2022a)]%
        {chen2022pali}
\bibfield{author}{\bibinfo{person}{Xi Chen}, \bibinfo{person}{Xiao Wang},
  \bibinfo{person}{Soravit Changpinyo}, \bibinfo{person}{AJ Piergiovanni},
  \bibinfo{person}{Piotr Padlewski}, \bibinfo{person}{Daniel Salz},
  \bibinfo{person}{Sebastian Goodman}, \bibinfo{person}{Adam Grycner},
  \bibinfo{person}{Basil Mustafa}, \bibinfo{person}{Lucas Beyer},
  {et~al\mbox{.}}} \bibinfo{year}{2022}\natexlab{a}.
\newblock \showarticletitle{Pali: A jointly-scaled multilingual language-image
  model}.
\newblock \bibinfo{journal}{\emph{arXiv preprint arXiv:2209.06794}}
  (\bibinfo{year}{2022}).
\newblock


\bibitem[Chen and Sun(2023)]%
        {chen2023LPCAM}
\bibfield{author}{\bibinfo{person}{Zhaozheng Chen} {and}
  \bibinfo{person}{Qianru Sun}.} \bibinfo{year}{2023}\natexlab{}.
\newblock \showarticletitle{Extracting Class Activation Maps from
  Non-Discriminative Features as well}. In
  \bibinfo{booktitle}{\emph{Proceedings of the IEEE/CVF Conference on Computer
  Vision and Pattern Recognition}}. \bibinfo{pages}{3135--3144}.
\newblock


\bibitem[Chen et~al\mbox{.}(2022b)]%
        {chen2022recam}
\bibfield{author}{\bibinfo{person}{Zhaozheng Chen}, \bibinfo{person}{Tan Wang},
  \bibinfo{person}{Xiongwei Wu}, \bibinfo{person}{Xian-Sheng Hua},
  \bibinfo{person}{Hanwang Zhang}, {and} \bibinfo{person}{Qianru Sun}.}
  \bibinfo{year}{2022}\natexlab{b}.
\newblock \showarticletitle{Class re-activation maps for weakly-supervised
  semantic segmentation}. In \bibinfo{booktitle}{\emph{Proceedings of the
  IEEE/CVF Conference on Computer Vision and Pattern Recognition}}.
  \bibinfo{pages}{969--978}.
\newblock


\bibitem[Deng et~al\mbox{.}(2009)]%
        {deng2009imagenet}
\bibfield{author}{\bibinfo{person}{Jia Deng}, \bibinfo{person}{Wei Dong},
  \bibinfo{person}{Richard Socher}, \bibinfo{person}{Li-Jia Li},
  \bibinfo{person}{Kai Li}, {and} \bibinfo{person}{Li Fei-Fei}.}
  \bibinfo{year}{2009}\natexlab{}.
\newblock \showarticletitle{Imagenet: A large-scale hierarchical image
  database}. In \bibinfo{booktitle}{\emph{2009 IEEE conference on computer
  vision and pattern recognition}}. Ieee, \bibinfo{pages}{248--255}.
\newblock


\bibitem[Everingham et~al\mbox{.}(2009)]%
        {everingham2009pascal}
\bibfield{author}{\bibinfo{person}{Mark Everingham}, \bibinfo{person}{Luc
  Van~Gool}, \bibinfo{person}{Christopher~KI Williams}, \bibinfo{person}{John
  Winn}, {and} \bibinfo{person}{Andrew Zisserman}.}
  \bibinfo{year}{2009}\natexlab{}.
\newblock \showarticletitle{The pascal visual object classes (voc) challenge}.
\newblock \bibinfo{journal}{\emph{International journal of computer vision}}
  \bibinfo{volume}{88} (\bibinfo{year}{2009}), \bibinfo{pages}{303--308}.
\newblock


\bibitem[He et~al\mbox{.}(2020)]%
        {kaiming2020moco}
\bibfield{author}{\bibinfo{person}{Kaiming He}, \bibinfo{person}{Haoqi Fan},
  \bibinfo{person}{Yuxin Wu}, \bibinfo{person}{Saining Xie}, {and}
  \bibinfo{person}{Ross Girshick}.} \bibinfo{year}{2020}\natexlab{}.
\newblock \showarticletitle{Momentum contrast for unsupervised visual
  representation learning}. In \bibinfo{booktitle}{\emph{Proceedings of the
  IEEE/CVF conference on computer vision and pattern recognition}}.
  \bibinfo{pages}{9729--9738}.
\newblock


\bibitem[He et~al\mbox{.}(2016)]%
        {he2016resnet}
\bibfield{author}{\bibinfo{person}{Kaiming He}, \bibinfo{person}{Xiangyu
  Zhang}, \bibinfo{person}{Shaoqing Ren}, {and} \bibinfo{person}{Jian Sun}.}
  \bibinfo{year}{2016}\natexlab{}.
\newblock \showarticletitle{Deep residual learning for image recognition}. In
  \bibinfo{booktitle}{\emph{Proceedings of the IEEE conference on computer
  vision and pattern recognition}}. \bibinfo{pages}{770--778}.
\newblock


\bibitem[Jia et~al\mbox{.}(2021)]%
        {jia2021ALIGN}
\bibfield{author}{\bibinfo{person}{Chao Jia}, \bibinfo{person}{Yinfei Yang},
  \bibinfo{person}{Ye Xia}, \bibinfo{person}{Yi-Ting Chen},
  \bibinfo{person}{Zarana Parekh}, \bibinfo{person}{Hieu Pham},
  \bibinfo{person}{Quoc Le}, \bibinfo{person}{Yun-Hsuan Sung},
  \bibinfo{person}{Zhen Li}, {and} \bibinfo{person}{Tom Duerig}.}
  \bibinfo{year}{2021}\natexlab{}.
\newblock \showarticletitle{Scaling up visual and vision-language
  representation learning with noisy text supervision}. In
  \bibinfo{booktitle}{\emph{International Conference on Machine Learning}}.
  PMLR, \bibinfo{pages}{4904--4916}.
\newblock


\bibitem[Jiang et~al\mbox{.}(2022)]%
        {jiang2022l2g}
\bibfield{author}{\bibinfo{person}{Peng-Tao Jiang}, \bibinfo{person}{Yuqi
  Yang}, \bibinfo{person}{Qibin Hou}, {and} \bibinfo{person}{Yunchao Wei}.}
  \bibinfo{year}{2022}\natexlab{}.
\newblock \showarticletitle{L2g: A simple local-to-global knowledge transfer
  framework for weakly supervised semantic segmentation}. In
  \bibinfo{booktitle}{\emph{Proceedings of the IEEE/CVF Conference on Computer
  Vision and Pattern Recognition}}. \bibinfo{pages}{16886--16896}.
\newblock


\bibitem[Khoreva et~al\mbox{.}(2017)]%
        {khoreva2017simple}
\bibfield{author}{\bibinfo{person}{Anna Khoreva}, \bibinfo{person}{Rodrigo
  Benenson}, \bibinfo{person}{Jan Hosang}, \bibinfo{person}{Matthias Hein},
  {and} \bibinfo{person}{Bernt Schiele}.} \bibinfo{year}{2017}\natexlab{}.
\newblock \showarticletitle{Simple does it: Weakly supervised instance and
  semantic segmentation}. In \bibinfo{booktitle}{\emph{Proceedings of the IEEE
  conference on computer vision and pattern recognition}}.
  \bibinfo{pages}{876--885}.
\newblock


\bibitem[Lee et~al\mbox{.}(2021a)]%
        {lee2021reducing}
\bibfield{author}{\bibinfo{person}{Jungbeom Lee}, \bibinfo{person}{Jooyoung
  Choi}, \bibinfo{person}{Jisoo Mok}, {and} \bibinfo{person}{Sungroh Yoon}.}
  \bibinfo{year}{2021}\natexlab{a}.
\newblock \showarticletitle{Reducing information bottleneck for weakly
  supervised semantic segmentation}.
\newblock \bibinfo{journal}{\emph{Advances in Neural Information Processing
  Systems}}  \bibinfo{volume}{34} (\bibinfo{year}{2021}),
  \bibinfo{pages}{27408--27421}.
\newblock


\bibitem[Lee et~al\mbox{.}(2021b)]%
        {lee2021anti}
\bibfield{author}{\bibinfo{person}{Jungbeom Lee}, \bibinfo{person}{Eunji Kim},
  {and} \bibinfo{person}{Sungroh Yoon}.} \bibinfo{year}{2021}\natexlab{b}.
\newblock \showarticletitle{Anti-adversarially manipulated attributions for
  weakly and semi-supervised semantic segmentation}. In
  \bibinfo{booktitle}{\emph{Proceedings of the IEEE/CVF Conference on Computer
  Vision and Pattern Recognition}}. \bibinfo{pages}{4071--4080}.
\newblock


\bibitem[Lee et~al\mbox{.}(2022b)]%
        {lee2022wood}
\bibfield{author}{\bibinfo{person}{Jungbeom Lee}, \bibinfo{person}{Seong~Joon
  Oh}, \bibinfo{person}{Sangdoo Yun}, \bibinfo{person}{Junsuk Choe},
  \bibinfo{person}{Eunji Kim}, {and} \bibinfo{person}{Sungroh Yoon}.}
  \bibinfo{year}{2022}\natexlab{b}.
\newblock \showarticletitle{Weakly supervised semantic segmentation using
  out-of-distribution data}. In \bibinfo{booktitle}{\emph{Proceedings of the
  IEEE/CVF Conference on Computer Vision and Pattern Recognition}}.
  \bibinfo{pages}{16897--16906}.
\newblock


\bibitem[Lee et~al\mbox{.}(2022a)]%
        {lee2022amn}
\bibfield{author}{\bibinfo{person}{Minhyun Lee}, \bibinfo{person}{Dongseob
  Kim}, {and} \bibinfo{person}{Hyunjung Shim}.}
  \bibinfo{year}{2022}\natexlab{a}.
\newblock \showarticletitle{Threshold matters in WSSS: manipulating the
  activation for the robust and accurate segmentation model against
  thresholds}. In \bibinfo{booktitle}{\emph{Proceedings of the IEEE/CVF
  Conference on Computer Vision and Pattern Recognition}}.
  \bibinfo{pages}{4330--4339}.
\newblock


\bibitem[Lee et~al\mbox{.}(2021c)]%
        {lee2021railroad}
\bibfield{author}{\bibinfo{person}{Seungho Lee}, \bibinfo{person}{Minhyun Lee},
  \bibinfo{person}{Jongwuk Lee}, {and} \bibinfo{person}{Hyunjung Shim}.}
  \bibinfo{year}{2021}\natexlab{c}.
\newblock \showarticletitle{Railroad is not a train: Saliency as pseudo-pixel
  supervision for weakly supervised semantic segmentation}. In
  \bibinfo{booktitle}{\emph{Proceedings of the IEEE/CVF conference on computer
  vision and pattern recognition}}. \bibinfo{pages}{5495--5505}.
\newblock


\bibitem[Li et~al\mbox{.}(2022b)]%
        {li2022blip}
\bibfield{author}{\bibinfo{person}{Junnan Li}, \bibinfo{person}{Dongxu Li},
  \bibinfo{person}{Caiming Xiong}, {and} \bibinfo{person}{Steven Hoi}.}
  \bibinfo{year}{2022}\natexlab{b}.
\newblock \showarticletitle{Blip: Bootstrapping language-image pre-training for
  unified vision-language understanding and generation}. In
  \bibinfo{booktitle}{\emph{International Conference on Machine Learning}}.
  PMLR, \bibinfo{pages}{12888--12900}.
\newblock


\bibitem[Li et~al\mbox{.}(2022a)]%
        {li2022urn}
\bibfield{author}{\bibinfo{person}{Yi Li}, \bibinfo{person}{Yiqun Duan},
  \bibinfo{person}{Zhanghui Kuang}, \bibinfo{person}{Yimin Chen},
  \bibinfo{person}{Wayne Zhang}, {and} \bibinfo{person}{Xiaomeng Li}.}
  \bibinfo{year}{2022}\natexlab{a}.
\newblock \showarticletitle{Uncertainty estimation via response scaling for
  pseudo-mask noise mitigation in weakly-supervised semantic segmentation}. In
  \bibinfo{booktitle}{\emph{Proceedings of the AAAI Conference on Artificial
  Intelligence}}, Vol.~\bibinfo{volume}{36}. \bibinfo{pages}{1447--1455}.
\newblock


\bibitem[Li et~al\mbox{.}(2021)]%
        {li2021declip}
\bibfield{author}{\bibinfo{person}{Yangguang Li}, \bibinfo{person}{Feng Liang},
  \bibinfo{person}{Lichen Zhao}, \bibinfo{person}{Yufeng Cui},
  \bibinfo{person}{Wanli Ouyang}, \bibinfo{person}{Jing Shao},
  \bibinfo{person}{Fengwei Yu}, {and} \bibinfo{person}{Junjie Yan}.}
  \bibinfo{year}{2021}\natexlab{}.
\newblock \showarticletitle{Supervision exists everywhere: A data efficient
  contrastive language-image pre-training paradigm}.
\newblock \bibinfo{journal}{\emph{arXiv preprint arXiv:2110.05208}}
  (\bibinfo{year}{2021}).
\newblock


\bibitem[Liang et~al\mbox{.}(2022)]%
        {liang2022ovseg}
\bibfield{author}{\bibinfo{person}{Feng Liang}, \bibinfo{person}{Bichen Wu},
  \bibinfo{person}{Xiaoliang Dai}, \bibinfo{person}{Kunpeng Li},
  \bibinfo{person}{Yinan Zhao}, \bibinfo{person}{Hang Zhang},
  \bibinfo{person}{Peizhao Zhang}, \bibinfo{person}{Peter Vajda}, {and}
  \bibinfo{person}{Diana Marculescu}.} \bibinfo{year}{2022}\natexlab{}.
\newblock \showarticletitle{Open-vocabulary semantic segmentation with
  mask-adapted clip}.
\newblock \bibinfo{journal}{\emph{arXiv preprint arXiv:2210.04150}}
  (\bibinfo{year}{2022}).
\newblock


\bibitem[Lin et~al\mbox{.}(2016)]%
        {lin2016scribblesup}
\bibfield{author}{\bibinfo{person}{Di Lin}, \bibinfo{person}{Jifeng Dai},
  \bibinfo{person}{Jiaya Jia}, \bibinfo{person}{Kaiming He}, {and}
  \bibinfo{person}{Jian Sun}.} \bibinfo{year}{2016}\natexlab{}.
\newblock \showarticletitle{Scribblesup: Scribble-supervised convolutional
  networks for semantic segmentation}. In \bibinfo{booktitle}{\emph{Proceedings
  of the IEEE conference on computer vision and pattern recognition}}.
  \bibinfo{pages}{3159--3167}.
\newblock


\bibitem[Lin et~al\mbox{.}(2014)]%
        {lin2014microsoft}
\bibfield{author}{\bibinfo{person}{Tsung-Yi Lin}, \bibinfo{person}{Michael
  Maire}, \bibinfo{person}{Serge Belongie}, \bibinfo{person}{James Hays},
  \bibinfo{person}{Pietro Perona}, \bibinfo{person}{Deva Ramanan},
  \bibinfo{person}{Piotr Doll{\'a}r}, {and} \bibinfo{person}{C~Lawrence
  Zitnick}.} \bibinfo{year}{2014}\natexlab{}.
\newblock \showarticletitle{Microsoft coco: Common objects in context}. In
  \bibinfo{booktitle}{\emph{Computer Vision--ECCV 2014: 13th European
  Conference, Zurich, Switzerland, September 6-12, 2014, Proceedings, Part V
  13}}. Springer, \bibinfo{pages}{740--755}.
\newblock


\bibitem[Lin et~al\mbox{.}(2022)]%
        {lin2022clipes}
\bibfield{author}{\bibinfo{person}{Yuqi Lin}, \bibinfo{person}{Minghao Chen},
  \bibinfo{person}{Wenxiao Wang}, \bibinfo{person}{Boxi Wu},
  \bibinfo{person}{Ke Li}, \bibinfo{person}{Binbin Lin},
  \bibinfo{person}{Haifeng Liu}, {and} \bibinfo{person}{Xiaofei He}.}
  \bibinfo{year}{2022}\natexlab{}.
\newblock \showarticletitle{CLIP is Also an Efficient Segmenter: A Text-Driven
  Approach for Weakly Supervised Semantic Segmentation}.
\newblock \bibinfo{journal}{\emph{arXiv preprint arXiv:2212.09506}}
  (\bibinfo{year}{2022}).
\newblock


\bibitem[Oord et~al\mbox{.}(2018)]%
        {oord2018infonce}
\bibfield{author}{\bibinfo{person}{Aaron van~den Oord}, \bibinfo{person}{Yazhe
  Li}, {and} \bibinfo{person}{Oriol Vinyals}.} \bibinfo{year}{2018}\natexlab{}.
\newblock \showarticletitle{Representation learning with contrastive predictive
  coding}.
\newblock \bibinfo{journal}{\emph{arXiv preprint arXiv:1807.03748}}
  (\bibinfo{year}{2018}).
\newblock


\bibitem[Pu et~al\mbox{.}(2018)]%
        {pu2018graphnet}
\bibfield{author}{\bibinfo{person}{Mengyang Pu}, \bibinfo{person}{Yaping
  Huang}, \bibinfo{person}{Qingji Guan}, {and} \bibinfo{person}{Qi Zou}.}
  \bibinfo{year}{2018}\natexlab{}.
\newblock \showarticletitle{GraphNet: Learning image pseudo annotations for
  weakly-supervised semantic segmentation}. In
  \bibinfo{booktitle}{\emph{Proceedings of the 26th ACM international
  conference on Multimedia}}. \bibinfo{pages}{483--491}.
\newblock


\bibitem[Radford et~al\mbox{.}(2021)]%
        {CLIP}
\bibfield{author}{\bibinfo{person}{Alec Radford}, \bibinfo{person}{Jong~Wook
  Kim}, \bibinfo{person}{Chris Hallacy}, \bibinfo{person}{Aditya Ramesh},
  \bibinfo{person}{Gabriel Goh}, \bibinfo{person}{Sandhini Agarwal},
  \bibinfo{person}{Girish Sastry}, \bibinfo{person}{Amanda Askell},
  \bibinfo{person}{Pamela Mishkin}, \bibinfo{person}{Jack Clark},
  {et~al\mbox{.}}} \bibinfo{year}{2021}\natexlab{}.
\newblock \showarticletitle{Learning transferable visual models from natural
  language supervision}. In \bibinfo{booktitle}{\emph{International Conference
  on Machine Learning}}. PMLR, \bibinfo{pages}{8748--8763}.
\newblock


\bibitem[Vernaza and Chandraker(2017)]%
        {vernaza2017learning}
\bibfield{author}{\bibinfo{person}{Paul Vernaza} {and}
  \bibinfo{person}{Manmohan Chandraker}.} \bibinfo{year}{2017}\natexlab{}.
\newblock \showarticletitle{Learning random-walk label propagation for
  weakly-supervised semantic segmentation}. In
  \bibinfo{booktitle}{\emph{Proceedings of the IEEE conference on computer
  vision and pattern recognition}}. \bibinfo{pages}{7158--7166}.
\newblock


\bibitem[Wang et~al\mbox{.}(2020)]%
        {wang2020seam}
\bibfield{author}{\bibinfo{person}{Yude Wang}, \bibinfo{person}{Jie Zhang},
  \bibinfo{person}{Meina Kan}, \bibinfo{person}{Shiguang Shan}, {and}
  \bibinfo{person}{Xilin Chen}.} \bibinfo{year}{2020}\natexlab{}.
\newblock \showarticletitle{Self-supervised equivariant attention mechanism for
  weakly supervised semantic segmentation}. In
  \bibinfo{booktitle}{\emph{Proceedings of the IEEE/CVF Conference on Computer
  Vision and Pattern Recognition}}. \bibinfo{pages}{12275--12284}.
\newblock


\bibitem[Wang et~al\mbox{.}(2021)]%
        {wang2021simvlm}
\bibfield{author}{\bibinfo{person}{Zirui Wang}, \bibinfo{person}{Jiahui Yu},
  \bibinfo{person}{Adams~Wei Yu}, \bibinfo{person}{Zihang Dai},
  \bibinfo{person}{Yulia Tsvetkov}, {and} \bibinfo{person}{Yuan Cao}.}
  \bibinfo{year}{2021}\natexlab{}.
\newblock \showarticletitle{Simvlm: Simple visual language model pretraining
  with weak supervision}.
\newblock \bibinfo{journal}{\emph{arXiv preprint arXiv:2108.10904}}
  (\bibinfo{year}{2021}).
\newblock


\bibitem[Wu et~al\mbox{.}(2019)]%
        {wu2019widerresnet}
\bibfield{author}{\bibinfo{person}{Zifeng Wu}, \bibinfo{person}{Chunhua Shen},
  {and} \bibinfo{person}{Anton Van Den~Hengel}.}
  \bibinfo{year}{2019}\natexlab{}.
\newblock \showarticletitle{Wider or deeper: Revisiting the resnet model for
  visual recognition}.
\newblock \bibinfo{journal}{\emph{Pattern Recognition}}  \bibinfo{volume}{90}
  (\bibinfo{year}{2019}), \bibinfo{pages}{119--133}.
\newblock


\bibitem[Xie et~al\mbox{.}(2022a)]%
        {xie2022clims}
\bibfield{author}{\bibinfo{person}{Jinheng Xie}, \bibinfo{person}{Xianxu Hou},
  \bibinfo{person}{Kai Ye}, {and} \bibinfo{person}{Linlin Shen}.}
  \bibinfo{year}{2022}\natexlab{a}.
\newblock \showarticletitle{CLIMS: cross language image matching for weakly
  supervised semantic segmentation}. In \bibinfo{booktitle}{\emph{Proceedings
  of the IEEE/CVF Conference on Computer Vision and Pattern Recognition}}.
  \bibinfo{pages}{4483--4492}.
\newblock


\bibitem[Xie et~al\mbox{.}(2021)]%
        {Xie_2021_ICCV}
\bibfield{author}{\bibinfo{person}{Jinheng Xie}, \bibinfo{person}{Cheng Luo},
  \bibinfo{person}{Xiangping Zhu}, \bibinfo{person}{Ziqi Jin},
  \bibinfo{person}{Weizeng Lu}, {and} \bibinfo{person}{Linlin Shen}.}
  \bibinfo{year}{2021}\natexlab{}.
\newblock \showarticletitle{Online Refinement of Low-Level Feature Based
  Activation Map for Weakly Supervised Object Localization}. In
  \bibinfo{booktitle}{\emph{Proceedings of the IEEE/CVF International
  Conference on Computer Vision (ICCV)}}. \bibinfo{pages}{132--141}.
\newblock


\bibitem[Xie et~al\mbox{.}(2022b)]%
        {C2AM}
\bibfield{author}{\bibinfo{person}{Jinheng Xie}, \bibinfo{person}{Jianfeng
  Xiang}, \bibinfo{person}{Junliang Chen}, \bibinfo{person}{Xianxu Hou},
  \bibinfo{person}{Xiaodong Zhao}, {and} \bibinfo{person}{Linlin Shen}.}
  \bibinfo{year}{2022}\natexlab{b}.
\newblock \showarticletitle{C2AM: Contrastive Learning of Class-Agnostic
  Activation Map for Weakly Supervised Object Localization and Semantic
  Segmentation}. In \bibinfo{booktitle}{\emph{Proceedings of the IEEE/CVF
  Conference on Computer Vision and Pattern Recognition (CVPR)}}.
  \bibinfo{pages}{989--998}.
\newblock


\bibitem[Xu et~al\mbox{.}(2022b)]%
        {xu2022boatinsky}
\bibfield{author}{\bibinfo{person}{Jianjun Xu}, \bibinfo{person}{Hongtao Xie},
  \bibinfo{person}{Hai Xu}, \bibinfo{person}{Yuxin Wang},
  \bibinfo{person}{Sun-ao Liu}, {and} \bibinfo{person}{Yongdong Zhang}.}
  \bibinfo{year}{2022}\natexlab{b}.
\newblock \showarticletitle{Boat in the Sky: Background Decoupling and
  Object-aware Pooling for Weakly Supervised Semantic Segmentation}. In
  \bibinfo{booktitle}{\emph{Proceedings of the 30th ACM International
  Conference on Multimedia}}. \bibinfo{pages}{5783--5792}.
\newblock


\bibitem[Xu et~al\mbox{.}(2022a)]%
        {xu2022mctformer}
\bibfield{author}{\bibinfo{person}{Lian Xu}, \bibinfo{person}{Wanli Ouyang},
  \bibinfo{person}{Mohammed Bennamoun}, \bibinfo{person}{Farid Boussaid}, {and}
  \bibinfo{person}{Dan Xu}.} \bibinfo{year}{2022}\natexlab{a}.
\newblock \showarticletitle{Multi-class token transformer for weakly supervised
  semantic segmentation}. In \bibinfo{booktitle}{\emph{Proceedings of the
  IEEE/CVF Conference on Computer Vision and Pattern Recognition}}.
  \bibinfo{pages}{4310--4319}.
\newblock


\bibitem[Xu et~al\mbox{.}(2023)]%
        {xu2023MMCST}
\bibfield{author}{\bibinfo{person}{Lian Xu}, \bibinfo{person}{Wanli Ouyang},
  \bibinfo{person}{Mohammed Bennamoun}, \bibinfo{person}{Farid Boussaid}, {and}
  \bibinfo{person}{Dan Xu}.} \bibinfo{year}{2023}\natexlab{}.
\newblock \showarticletitle{Learning Multi-Modal Class-Specific Tokens for
  Weakly Supervised Dense Object Localization}. In
  \bibinfo{booktitle}{\emph{Proceedings of the IEEE/CVF Conference on Computer
  Vision and Pattern Recognition}}. \bibinfo{pages}{19596--19605}.
\newblock


\bibitem[Zhang et~al\mbox{.}(2020)]%
        {zhang2020conta}
\bibfield{author}{\bibinfo{person}{Dong Zhang}, \bibinfo{person}{Hanwang
  Zhang}, \bibinfo{person}{Jinhui Tang}, \bibinfo{person}{Xian-Sheng Hua},
  {and} \bibinfo{person}{Qianru Sun}.} \bibinfo{year}{2020}\natexlab{}.
\newblock \showarticletitle{Causal intervention for weakly-supervised semantic
  segmentation}.
\newblock \bibinfo{journal}{\emph{Advances in Neural Information Processing
  Systems}}  \bibinfo{volume}{33} (\bibinfo{year}{2020}),
  \bibinfo{pages}{655--666}.
\newblock


\bibitem[Zhang et~al\mbox{.}(2022)]%
        {zhang2022multi}
\bibfield{author}{\bibinfo{person}{Meijie Zhang}, \bibinfo{person}{Jianwu Li},
  {and} \bibinfo{person}{Tianfei Zhou}.} \bibinfo{year}{2022}\natexlab{}.
\newblock \showarticletitle{Multi-Granular Semantic Mining for Weakly
  Supervised Semantic Segmentation}. In \bibinfo{booktitle}{\emph{Proceedings
  of the 30th ACM International Conference on Multimedia}}.
  \bibinfo{pages}{6019--6028}.
\newblock


\bibitem[Zhang et~al\mbox{.}(2021)]%
        {zhang2021adaptive}
\bibfield{author}{\bibinfo{person}{Xiangrong Zhang}, \bibinfo{person}{Zelin
  Peng}, \bibinfo{person}{Peng Zhu}, \bibinfo{person}{Tianyang Zhang},
  \bibinfo{person}{Chen Li}, \bibinfo{person}{Huiyu Zhou}, {and}
  \bibinfo{person}{Licheng Jiao}.} \bibinfo{year}{2021}\natexlab{}.
\newblock \showarticletitle{Adaptive affinity loss and erroneous pseudo-label
  refinement for weakly supervised semantic segmentation}. In
  \bibinfo{booktitle}{\emph{Proceedings of the 29th ACM International
  Conference on Multimedia}}. \bibinfo{pages}{5463--5472}.
\newblock


\bibitem[Zhou et~al\mbox{.}(2016)]%
        {CAM}
\bibfield{author}{\bibinfo{person}{Bolei Zhou}, \bibinfo{person}{Aditya
  Khosla}, \bibinfo{person}{Agata Lapedriza}, \bibinfo{person}{Aude Oliva},
  {and} \bibinfo{person}{Antonio Torralba}.} \bibinfo{year}{2016}\natexlab{}.
\newblock \showarticletitle{Learning deep features for discriminative
  localization}. In \bibinfo{booktitle}{\emph{Proceedings of the IEEE
  conference on computer vision and pattern recognition}}.
  \bibinfo{pages}{2921--2929}.
\newblock


\bibitem[Zhou et~al\mbox{.}(2022)]%
        {zhou2022regional}
\bibfield{author}{\bibinfo{person}{Tianfei Zhou}, \bibinfo{person}{Meijie
  Zhang}, \bibinfo{person}{Fang Zhao}, {and} \bibinfo{person}{Jianwu Li}.}
  \bibinfo{year}{2022}\natexlab{}.
\newblock \showarticletitle{Regional semantic contrast and aggregation for
  weakly supervised semantic segmentation}. In
  \bibinfo{booktitle}{\emph{Proceedings of the IEEE/CVF Conference on Computer
  Vision and Pattern Recognition}}. \bibinfo{pages}{4299--4309}.
\newblock


\end{thebibliography}

\clearpage
\appendix

\section{Questions templates}

We defined 10 background questions templates and 6 foreground question templates for the Question-Answer Prompt Engineering procedure, as shown in Table \ref{tab:apdx-questions}.

\section{Filter Ratio Selection}

\begin{figure}[h]
  \centering
  \includegraphics[width=0.8\linewidth]{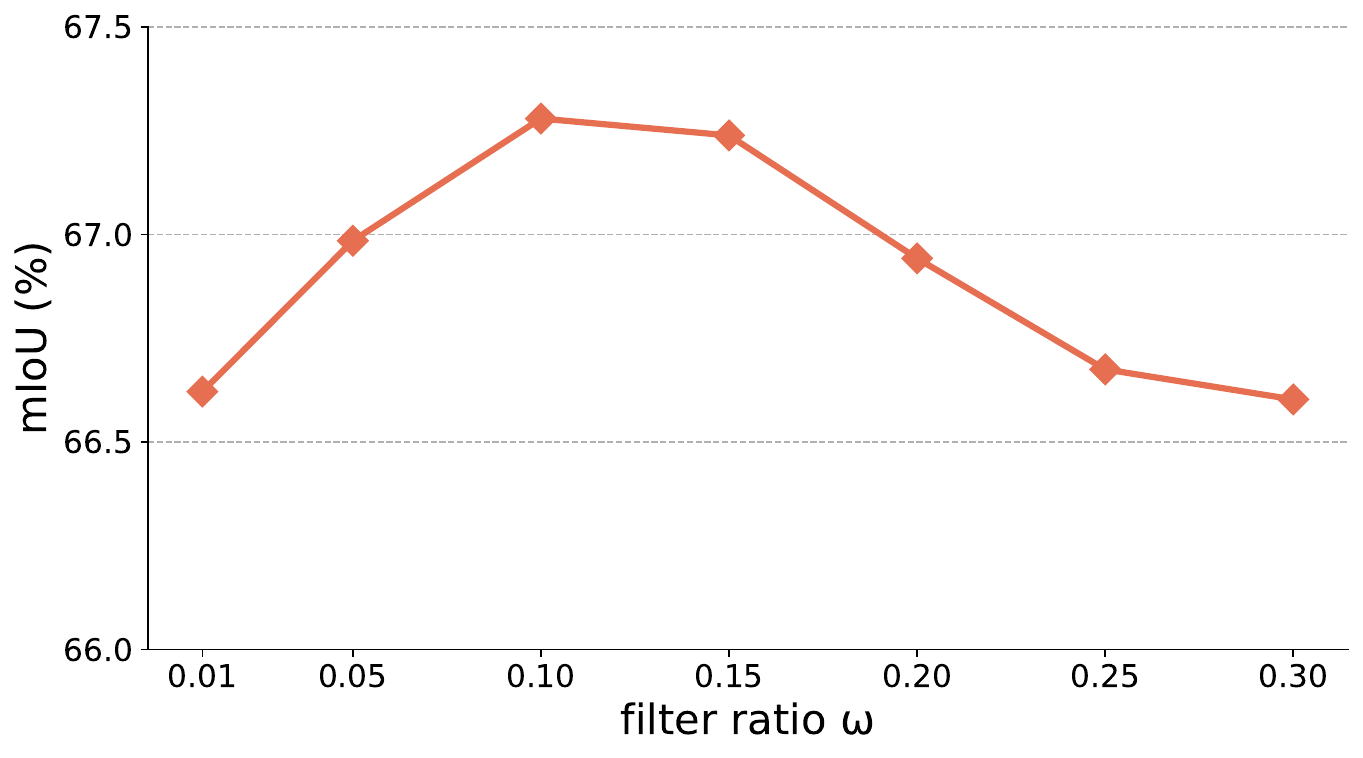}
  \caption{Impact of filter ratio on PASCAL VOC 2012.}
  \Description{None}
  \label{fig:apdx-alpha}
\end{figure}

In the foreground adaptive thresholding module, the filter ratio $\omega$ controls the extent to which the foreground region is erased (see Eq.\ref{eqn-mask-theta} in the main paper).
To investigate the impact of different values of $\omega$ on the performance of the module on VOC dataset, we plot the mIoU of the generated CAMs for different values of $\omega$ in Figure \ref{fig:apdx-alpha}.
The results show that there is an optimal value of $\omega$ that leads to the highest mIoU. 
We observe that the mIoU is highest when $\omega$ is set to 0.1, with a value of 67.3. As $\omega$ increases or decreases from this optimal value, the mIoU decreases.
This suggests that setting the filter ratio to an appropriate value can improve the performance of the foreground adaptive thresholding module.

\section{Training Details of Segmentation Network}

We supplement the details in the training process of the segmentation network.
For DeepLabV2, we follow \cite{xie2022clims, chen2022recam} cropped each training image to the size of $321 \times 321$.
The model is trained for 30k iterations with a batch size of 5 for VOC dataset and 100k iterations with a batch size of 10 for COCO dataset.
The learning rate and weight decay was set as 2.5e-4 and 5e-4, respectively.
For DeepLabV1, we cropped each training image to the size of $512 \times 512$ and train the model for 30k iterations with a batch size of 6. The learning rate was set as 8e-4.

\begin{table}[b]
  \caption{Full question templates for QAPE.}
  \label{tab:apdx-questions}
  \resizebox{\linewidth}{!}{
  \begin{tabular}{ccl}
    \toprule
    \multicolumn{2}{l}{Question Type} & Question Templates \\
    \midrule
    \ \multirow{10}*{BG}
    & \multirow{7}*{Surrounding Objects}
    & "\textit{What is above the \{class\}?}"\\
    \ & & "\textit{What is under the \{class\}?}" \\
    \ & & "\textit{What is behind the \{class\}?}"   \\
    \ & & "\textit{What is around the \{class\}?}"  \\
    \ & & "\textit{What is next to the \{class\}?}"  \\
    \ & & "\textit{What is the left side of \{class\}?}"   \\
    \ & & "\textit{What is the right side of \{class\}?}"  \\
    \cmidrule(r){2-3}
    \ & \multirow{3}*{Scene}
    & "\textit{What scene is the \{class\} in?}" \\
    \ & & "\textit{What enviroment is the \{class\} in?}" \\
    \ & & "\textit{What place is the \{class\} in?}" \\
    \cmidrule(r){1-3}
    \ \multirow{6}*{FG}
    & \multirow{2}*{Fine-grained Names}
    & "\textit{What kind of \{class\} is in the photo?}" \\
    \ & &"\textit{What type of \{class\} is in the photo?}" \\
    \cmidrule(r){2-3}
    \ & \multirow{4}*{Object Aliases}
    & "\textit{What is this \{class\} also called?}"  \\
    \ & & "\textit{What is this \{class\} usually called?}"  \\
    \ & & "\textit{What is another word for this \{class\}?}"  \\
    \ & & "\textit{What is another name for this \{class\}?}"  \\
  \bottomrule
\end{tabular}
}
\end{table}

\section{More Visualization}

\subsection{Class Activation Maps}

\begin{figure*}[t]
  \centering
  \includegraphics[width=\textwidth]{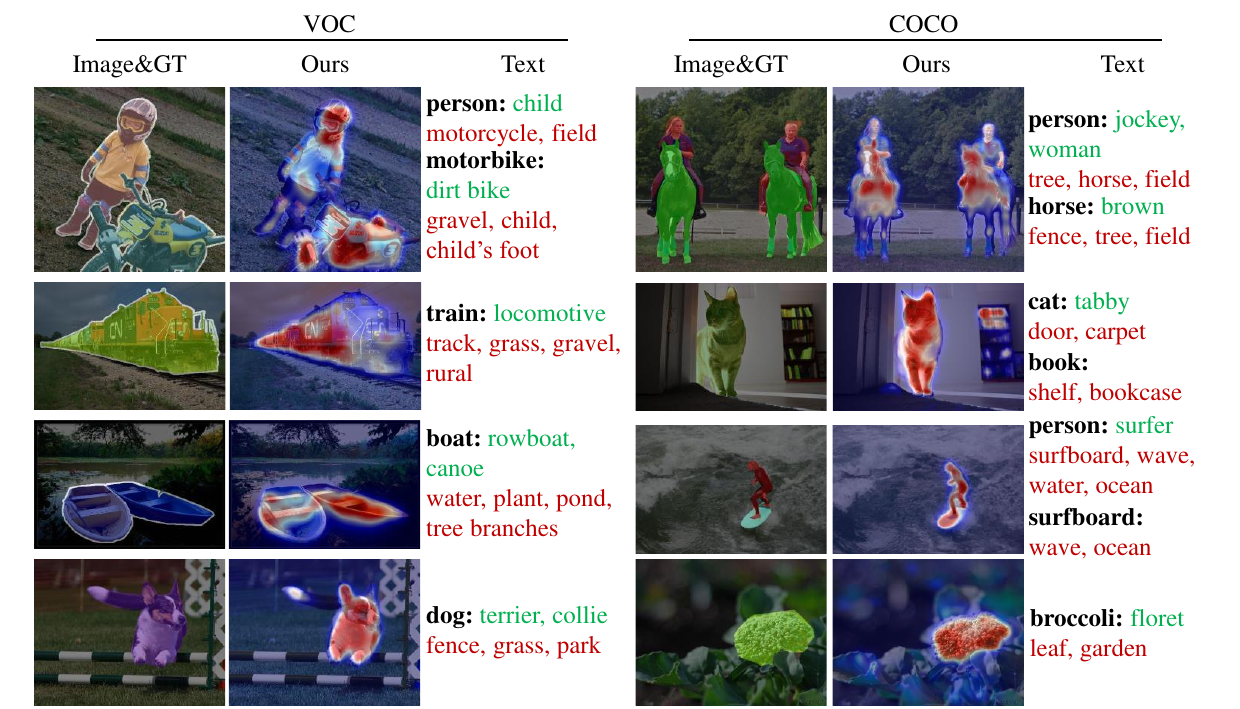}
  \caption{More generated CAMs visualizations on PASCAL VOC 2012 and MS COCO 2014 dataset.}
  \Description{None}
  \label{fig:apdx-vis_cam}
\end{figure*}

We present additional visualization of the class activation maps (CAMs) generated by our QA-CLIMS, along with the corresponding text labels generated by the QAPE module.
The visualizations are depicated in Figure \ref{fig:apdx-vis_cam}.

\subsection{Semantic Segmentation Resutls}

\begin{figure*}[t]
  \centering
  \includegraphics[width=\textwidth]{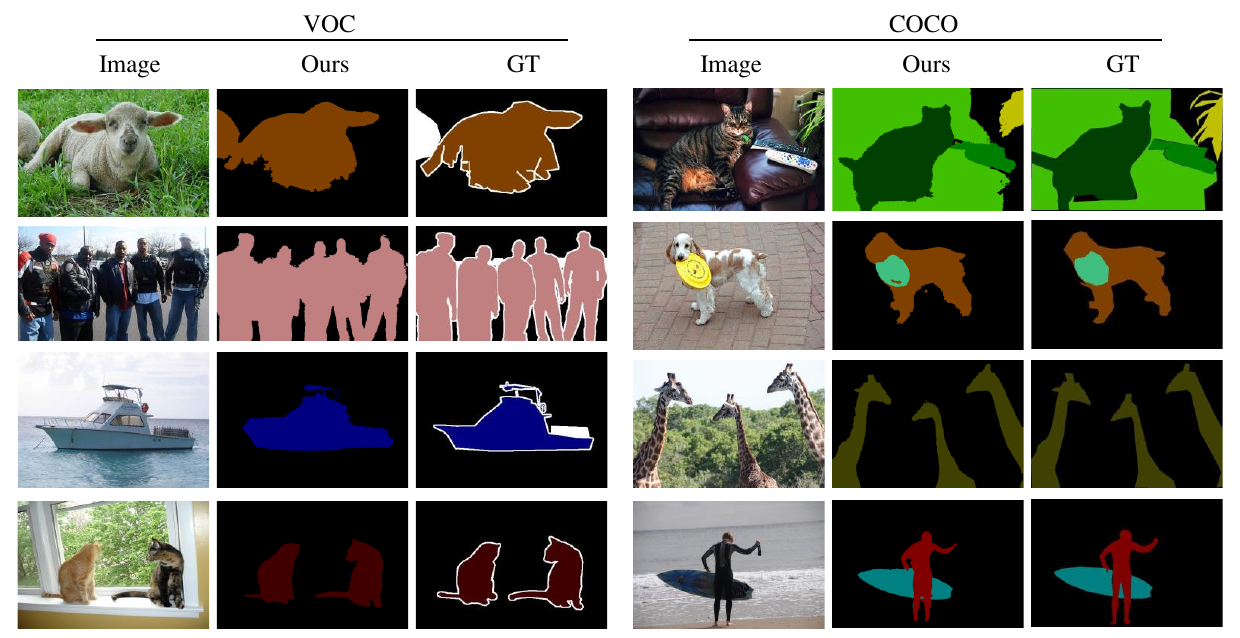}
  \caption{Semantic segmentation results on PASCAL VOC 2012 \textit{val} set and MS COCO 2014 \textit{val} set.}
  \Description{None}
  \label{fig:apdx-vis_seg}
\end{figure*}

Figure \ref{fig:apdx-vis_seg} showcases the visualization results of WSSS.

\end{document}